\documentclass[sigconf]{acmart}
\usepackage{epstopdf}
\usepackage{amsmath}
\usepackage{multirow}
\usepackage{colortbl}
\usepackage{subcaption}
\usepackage{bm}
\usepackage{amssymb}
\usepackage{verbatim}

\usepackage{booktabs} % For formal tables
\usepackage{enumitem}
\usepackage{amsfonts}
\begin{document}

\copyrightyear{2018} 
\acmYear{2018} 
\setcopyright{acmcopyright}
\acmConference[WSDM 2018]{WSDM 2018: The Eleventh ACM International Conference on Web Search and Data Mining }{February 5--9, 2018}{Marina Del Rey, CA, USA}
%\acmBooktitle{WSDM 2018: WSDM 2018: The Eleventh ACM International Conference on Web Search and Data Mining , February 5--9, 2018, Marina Del Rey, CA, USA}
\acmBooktitle{WSDM 2018: The Eleventh ACM International Conference on Web Search and Data Mining, February 5--9, 2018, Marina Del Rey, CA, USA}
\acmPrice{15.00}
\acmDOI{10.1145/3159652.3159703}
\acmISBN{978-1-4503-5581-0/18/02}

\fancyhead{}

%set the vertical space above and below the equation.
%\setlength{\abovedisplayskip}{3pt}
%\setlength{\belowdisplayskip}{3pt}
\title{Dynamic Word Embeddings for Evolving Semantic Discovery}

%\author{Zijun Yao\dagger, Yifan Sun\ddagger, Weicong Ding\S, Nikhil Rao\S, Hui Xiong\dagger}
%\affiliation{%
%  \institution{\dagger Rutgers University\\
%  Technicolor Research, yifan.sun@technicolor.com\\
%  Amazon}
%}
%\email{a}

\author{Zijun Yao}
\affiliation{%
  \institution{Rutgers University}
  }
\email{zijun.yao@rutgers.edu}
%}
\author{Yifan Sun}
\affiliation{%
  \institution{Technicolor Research}
  }
\email{yifan.sun@technicolor.com}
\author{Weicong Ding}
\affiliation{%
  \institution{Amazon}
}
\email{20008005dwc@gmail.com}
\author{Nikhil Rao}
\affiliation{%
  \institution{Amazon}
}
\email{nikhilrao86@gmail.com}
\author{Hui Xiong}
\affiliation{%
  \institution{Rutgers University}
}
\email{hxiong@rutgers.edu}
\newcommand{\eg}{\textit{e.g.}}
\newcommand{\ie}{\textit{i.e.}}
\newcommand{\etc}{\textit{etc.}}
\begin{abstract}
	Word evolution refers to the changing meanings and associations of words throughout time, as a byproduct of human language evolution.
By studying word evolution, we can infer social trends and language constructs over different periods of human history.
However, traditional techniques such as word representation learning do not adequately capture the evolving language structure and vocabulary. 
In this paper, we develop a dynamic statistical model to learn time-aware word vector representation. 
We propose a model that simultaneously learns time-aware embeddings and solves the resulting ``alignment problem''. 
This model is trained on  a crawled NYTimes dataset. Additionally, we develop multiple intuitive evaluation strategies of temporal word embeddings. 
Our qualitative and quantitative tests indicate that  our method not only reliably captures this evolution over time, but also consistently outperforms state-of-the-art temporal embedding approaches on both semantic accuracy and alignment quality.
\end{abstract}

%\keywords{Dynamic word embeddings; word semantic analysis}

\maketitle

%\vspace{-0.3cm}
\section{Introduction}
Human language is an evolving construct, with word semantic associations changing over time.
For example, \texttt{apple} which was traditionally only associated with fruits, is now also associated with a technology company. Similarly, the association of names of famous personalities  (\eg, \texttt{trump}) changes with a change in their roles. For this reason, understanding and tracking word evolution is useful for time-aware knowledge extraction tasks (\eg, public sentiment analysis), and other applications in text mining. 
To this end, we aim to learn word embeddings with a temporal bent, for capturing time-aware meanings of words.
%trained on a large-scale time-stamped corpora such as newspapers. 

Word embeddings aim to represent words with low-dimensional vectors, where words with similar semantics are geometrically closer (\eg~ \texttt{red} and \texttt{blue} are closer than \texttt{red} and \texttt{squirrel}).
Classic word embedding techniques started in the 90s and relied on statistical approaches \cite{HAL1996,lsa}.
Later, neural network approaches \cite{bengioneural}, as well as recent advances such as word2vec \cite{mikolov2013efficient,mikolov2013distributed} and GloVE \cite{glove} have greatly improved the performance of word representation learning.
However, these techniques usually do not consider temporal factors, and assume that the word is static across time.

In this paper, we are interested in computing time-aware embedding of words.
Specifically, each word in a different time frame (\eg, years) is represented by a different vector. From these embeddings, we have a better notion of ``distance'' (the cosine distance between word embedding vectors), and  by looking at word ``neighborhoods'' (defined through this distance),  we 
can better understand word associations as well as word meanings, as they evolve over time.

For instance, by locating the embeddings of personality names such as \texttt{trump}'s closest words, we can see that he can be associated with the trajectory : \texttt{real estate} $\rightarrow$ \texttt{television} $\rightarrow$ \texttt{republican}. Similarly, the trajectory of \texttt{apple} travels from the neighborhood of   \texttt{strawberry, mango} to that of \texttt{iphone, ipad}.

A key practical issue of learning different word embeddings for different time periods is alignment. Specifically, most cost functions for training are invariant to rotations, as a byproduct, the learned embeddings across time may not be placed in the same latent space.
We call this the \textit{alignment} problem, which is an issue in general if embeddings are learned independently for each time slice. 

Unlike traditional methods, literature on learning temporal word embedding is relatively short: \cite{zhang2016past, liao2016analysing, kulkarni2015statistically,hamilton2016diachronic}.
In general, the approaches in these works follow a similar two-step pattern: first compute static word embeddings in each time slice separately, then find a way to align the word embeddings across time slices.
To achieve alignment,
\cite{kulkarni2015statistically} finds a linear transformation of words between any two time slices by solving a $d$-dimensional least squares problem of $k$ nearest neighbor words (where $d$ is the embedding dimension).
Additionally, \cite{zhang2016past} also use the linear transformation approach between a base and target time slices, and computes the linear transformation using anchor words, which does not change meaning between the two time slices. This method requires the prior knowledge of words that are in fact static, which involves additional expert supervision. 
Finally, \cite{hamilton2016diachronic} imposes the transformation to be orthogonal, and solves a $d$-dimensional Procrustes problem between every two adjacent time slices. 
%However, these approaches do not achieve enough satisfied performance by separating the alignment phase from representation learning phase.

Our main novelty in this paper is to learn the word embeddings across time jointly, thus obviating the need to solve a separate alignment problem. Specifically, we propose to learn temporal embeddings in all time slices concurrently, and apply regularization terms to smooth embedding changes across time.
There are three main advantages to this proposed joint modeling approach:
\begin{itemize}
\item First, this can be seen as an improvement over traditional, ``single-time'' methods such as word2vec.
\item Second, our experimental results suggest that enforcing alignment through regularization yields better results than two-step methods.
\item Third, we can share information across time slices for the majority of vocabulary. As a result, our method is robust against data sparsity -- we can afford to have time slices where some words are rarely present, or even missing. This is a crucial practical advantage offered by our method. 
\end{itemize}

%\deleted[remark={since we aren't explaining this, it reads like an empty claim}]{In fact, 2-stage methods can be seen to learn a suboptimal solution to our optimization problem.}

%First, being different from applying post-learning alignment on classic word2vec model results, this approach is novel for improving word2vec model itself for handling temporal word embedding learning.
%Second, as time-to-time constrain is functioning in each embedding update, the alignment of embeddings is built into the optimization process and involves embeddings across time slices simultaneously. Approved by experimental results, the quality of alignment is better than two-step approaches.
%Third, in proposed model setting, we can share information across time slices for the majority of vocabulary. As a result, our method is more robust against data sparsity such as word occurrence absence in some time slice.
%In fact, previous two-step (learning-alignment) solutions optimally learn segments of our model independently, and thus can be viewed as suboptimal solutions of our model.

Since our model requires embeddings across all time slices to be learned at the same time, it can be computationally challenging. To mitigate this,  we employ a block coordinate descent method which can handle large vocabulary sizes through decomposition.
%, allowing for scalability. 
%\textcolor{red}{necessary footnote? Not really, can remove it. } \footnote{Though the method will of course be slower than learning embeddings independently for each time, for the latter methods we need to factor in the time required to solve the alignment problem.}

%In addition, Coordinate descent based methods have been shown to be faster then SGD based approaches, when the data can be made to fit in memory \cite{something}

In experimental study, we learn temporal embeddings of words from The New York Times articles between 1990 and 2016.
%First, we offer a new training corpus of around 100,000 major New York Times articles from 1990 to 2016. 
In contrast, previous temporal word embedding works have focused on time-stamped novels and magazine collections (such as Google N-Gram and COHA). However, news corpora are naturally advantageous to studying language evolution through the lens of current events. In addition, it allows us to work with a corpus that maintains consistency in narrative style and grammar, as opposed to Facebook and Twitter posts. For evaluating our embeddings, we develop both qualitative and quantitative metrics. 

%For evaluating temporal embeddings, we develop both qualitative and quantitative methods, although it has traditionally been difficult because of the inherent subjectiveness in evaluating word semantics, and a dearth of labeled data.
% which are traditionally difficult for evaluation by being a unsupervised task. 

\textbf{Qualitatively}, we illustrate the advantages of temporal embeddings for evolving semantics discovery by 1) plotting word vector trajectories to find evolving meanings and associations, 2) using alignment through time to identify associated words across time, and 3) looking at norms as a representative of concept popularity.

\textbf{Quantitatively}, we first use semantic topics extracted from Sections (\eg, \textit{Technology}, \textit{World News}) of news articles as ground truth to evaluate the semantic accuracy of temporal embeddings.
Additionally, we provide two testsets to evaluate cross-time alignment quality: one consists of known changing roles (\eg, U.S. presidents), determined objectively, and one of concept replacements (\eg, compact disk to mp3), determined more subjectively.
%consist of known changing roles (\eg, U.S presidents) and concept replacement (e.g., disk to mp3) respectively to evaluate cross-time alignment quality.
%These test sets can be used to evaluate temporal embeddings in general.
The experimental results show the effectiveness of our proposed model and demonstrate substantial improvements against baseline methods.

%In summary, the contributions of this work are as follows:
%\begin{itemize} [leftmargin=*]
%\item We present a unified dynamic model which incorporates embedding alignment among different time slices into the embedding learning process. Our model provides accurate word embedding, with high quality cross-time alignment, and is robust to data sparsity.
%\item We implement our proposed model on New York Times articles across 27 years. By visualizing the embeddings, we find interesting insights to semantic evaluations.
%%We carry out interesting qualitative analysis and visualization for showing the evolution of human knowledge. 
%\item We develop quantitative tests for evaluating our temporal word embeddings against state-of-the-art baselines. 
%\end{itemize}

The rest of this paper is organized as follows. 
In Section \ref{sec:method}, we present the proposed model for temporal embedding learning,
and in Section \ref{sec:bcd}, we describe a scalable algorithm to train it.
In Section \ref{sec:eval}, we describe the news corpus dataset and setup details of experiments. 
We perform qualitative evaluations in Section \ref{sec:qualitative}. Finally, we quantitatively compare our embeddings against other state-of-the-art temporal embeddings in Section \ref{sec:expt}.
%in two aspects: semantic similarity, and alignment quality.

%%%%%%%%%%%%%%%%%%%%%%%%%%%%
%%%%%%%%%%%%%%%%%%%%%%%%%%%%
%%%%%%%%%%%%%%%%%%%%%%%%%%%%
%%%%%%%%%%%%%%%%%%%%%%%%%%%%
%\vspace{-0.2cm}
\section{Methodology}
\label{sec:method}

We now set up our temporal word embedding model. We consider a text corpus collected across time. These kinds of corpora such as news article collections with published dates or social media discussion with time-stamps are ubiquitous. Formally, we denote by $\mathcal D = (\mathcal D_1,\hdots, \mathcal D_T)$ our text corpus where each $\mathcal D_t$, $t=1,\hdots T$, is the corpus of all documents in the $t$-th time slice. Without loss of generality, we assume the time slices are ordered chronologically.  The length of these time slices can be on the order of months, years, or decades. Moreover, the length of all the time slices could be different.  
We consider an overall vocabulary $\mathcal V = \{w_1, \hdots, w_V \}$ of size $V$. We note that the vocabulary $\mathcal V$ consists of words present in the corpus at any point in time, and thus it is possible for some  $w\in \mathcal V$ to not appear at all in some $\mathcal D_t$. This includes emerging words and dying words that are typical in real-world news corpora. 
%We first set up notations for the remainder of this paper. 
%
% We consider a time-tagged text corpus collected from $T$ different time slices $\mathcal D = (\mathcal D_1,\hdots, \mathcal D_T)$. $D_t$
%For a given corpus, the \emph{vocabulary} is an ordered unique set of $V$ relevant words $\mathcal V = (w_1, \hdots, w_V)$.
%The corpus is then divided into $T$ time slices $\mathcal D = (\mathcal D_1,\hdots, \mathcal D_T)$ where each $D_t$, $t=1,\hdots T$ is the ordered set of words corresponding to the concatenation of all documents in time slice $t$.
%The length of these time slices can be anything (months, years, even decades). 

Given such a time-tagged corpus, our goal is to find a dense, low-dimensional vector representation $u_w(t)\in\mathbb{R}^{d},~\ d \ll V$ for each word $w\in\mathcal{V}$ and each time period $t=1,\ldots, T$. We denote by $u_w$ the static embedding for word $w$ (for example, learned via word2vec), and $d$ is the embedding dimension (typically $50\leq d \leq 200$). Compactly, we denote by $U(t)$ (of size $V \times d$) the embedding matrix of all words whose $i$-th row corresponds to the embedding vector of $i$-th word $u_{w_i}(t)$. 

\vspace{-0.2cm}
\subsection{Time-agnostic word embeddings}
A fundamental observation in static word embedding literature is that semantically similar words often have similar neighboring words in a corpus \cite{firth1957synopsis}. 
This is the idea behind learning dense low-dimensional word representations both traditionally \cite{HAL1996,lsa,bengioneural} and recently \cite{mikolov2013efficient, glove}.
In several of these methods, the neighboring structure is captured by the frequencies by which pairs of words co-occur within a small local window. 

%In this paper  we adopt the approach of word2vec  \cite{mikolov2013efficient} and GloVE \cite{glove} and our embedding is also based on the same local co-occurrence structure.  

We compute the $V\times V$ pointwise mutual information (PMI) matrix specific to a corpus $\mathcal D$, whose $w,c$-th entry is:
\begin{equation}\label{eq:pmi}
\text{PMI}(\mathcal D, L)_{w,c} =  \log \left( \frac{\#(w,c)\cdot|\mathcal D|}{\#(w)\cdot\#(c)} \right),
\end{equation}
where $\#(w,c)$ counts the number of times that words $w$ and $c$ co-occur within a window of size $L$ in corpus $\mathcal D$ , and $\#(w)$, $\#(c)$ counts the number of occurrences of words $w$ and $c$ in $\mathcal D$. $\vert \mathcal{D}\vert$ is total number of word tokens in the corpus. $L$ is typically around $5$ to $10$; we set $L=5$ throughout this paper. 

%
%%%%%%%%%%%%%%
The key idea behind both word2vec \cite{mikolov2013efficient} and GloVE \cite{glove} is to find embedding vectors $u_w$ and $u_c$ such that for any $w, c$ combination, 
\begin{equation}
u_w^Tu_c \approx \text{PMI}(\mathcal D, L)_{w,c},
\end{equation}
where each $u_w$ has length $d\ll V$.
While both \cite{mikolov2013efficient} and \cite{glove} offer highly scalable algorithms such as negative sampling to do this implicitly, \cite{levy2014neural} shows that these are equivalent to low-rank factorization of $\text{PMI}(\mathcal{D},L)$ \footnote{with a constant shift that can be zero.}. Our approach is primarily motivated by this observation. We note that though the PMI matrices are of size $V\times V$, in real-world datasets it is typically sparse as observed in \cite{glove}, for which efficient factorization methods exist \cite{yu2012scalable}. 

%our model is primarily motivated by the observation of \cite{levy2014neural}, which showed that these are equivalent to factorizing the PMI matrix (+ a constant) directly.
%

\vspace{-0.2cm}
\subsection{Temporal word embeddings}
A natural extension of the static word embedding intuition is to use this matrix factorization technique on each time slice $\mathcal D_t$ separately. Specifically, for each time slice $t$, we define the $w,c$-th entry of \emph{positive} PMI matrix ($\text{PPMI}(t,L)$) as
\footnote{We consider the PPMI rather than the PMI because when $\frac{\#(w,c)\cdot|\mathcal D|}{\#(w)\cdot\#(c)} $ is very small, taking the log results in large negative values and is thus extremely unstable. Since for most significantly related pairs $w$ and $c$ the log argument is $>1$, thresholding it in this way will not affect the solution significantly, but will offer much better numerical stability. This approach is not unique to us; \cite{levy2014neural} also factorizes the PPMI.}
\begin{equation}
\label{ppmidefinition}
\text{PPMI}(t, L)_{w,c} = \max\{\text{PMI}(\mathcal D_t, L)_{w,c},0\}. := Y(t).
\end{equation}
The temporal word embeddings $U(t)$ must satisfy
%are learned by factorizing the $V\times V$ matrices $\text{PPMI}(t, L)$ independently:
% The  $d$-dimensional temporal word vectors are denoted as $u_{w}(t)$ for all $w\in \mathcal V$ and $t = 1,\hdots, T$, and should satisfy
%\[
%u_w(t)^Tu_c(t) \approx \text{PPMI}(w,c,t),
%\]
% or, more compactly,
\begin{equation}\label{e-model1}
U(t)U(t)^T \approx \text{PPMI}(t, L).
\end{equation}
One way to find such $U(t)$ is for each  $t$, factorizing $\text{PPMI}(t,L)$ by either using an eigenvalue method or solving  a matrix factorization problem iteratively.  

\noindent\textbf{The Alignment Problem:}~
%The most obvious way to determine $U(t)$ satisfying \eqref{e-model1} is for each $t$, factorize $Y(t) = PPMI(t,L)$ using either an eigenvalue method or solving  a matrix factorization problem iteratively. However, 
Imposing \eqref{e-model1} is not sufficient for a unique embedding, since the solutions are invariant under rotation; that is, 
for any $d\times d$ orthogonal matrix $R$, we have the embedding $\widehat{U}(t) =U(t)  R$ , the approximation error in \eqref{e-model1} is the same since
%\[
%\|u_w(t)-u_c(t)\|_2 = \|Ru_w(t)= Ru_c(t)\|.
%\]
\[
\widehat{U}(t) \widehat{U}(t)^T =  U(t)RR^TU(t)^T = U(t)U(t)^T.
\]
%
% We note that the optimal embedding in Eq.~\eqref{eq:mf_static} or Eq.~\eqref{eq:skipgram} is not unique. Indeed, any ortho-normal transformation on an optimal embedding ($U^{\prime} = U R$ where $R$ is $d\times d$ matrix and $RR^{\top}= I$ ) is also optimal. 
For this reason, it is important to enforce \emph{alignment}; if word $w$ did not semantically shift from $t$ to $t+1$, then we additionally require $u_w(t)\approx u_w(t+1)$. 
To do this, \cite{kulkarni2015statistically, hamilton2016diachronic} propose two-step procedures; first, they factorize each $Y(t)$ separately, and afterwards enforce alignment using  local linear mapping \cite{kulkarni2015statistically}  or solving an orthogonal procrustes problem \cite{hamilton2016diachronic}. 
Note that in these methods, aligning $U(t)$ to $U(t')$ assumes that 
we desire $U(t) \approx U(t')$. 
%
%; in fact, we only wish this for the columns of $U(t)$, $U(t')$ that have not semantically shifted. 
If we only pick $t' = t+1$ (as done in \cite{hamilton2016diachronic}), this assumption is reasonable because between any two years, only a few words experience semantic shift, emergence, or death.  
However, this becomes problematic if $U(t)$ was a result of a time period with extremely sparse data (and hence poorly learned); all subsequent year embeddings and previous year embeddings will be poorly aligned.

\vspace{-0.2cm}
\subsection{Our model}
%We propose a novel temporal word embedding objective that account for both the linguistic similarity within each time period and the evolution of word semantics across time horizon. 
%Geometrically, we enforce the trajectory of each word $\{ u(t) \} t=1,\ldots,T$ to be {\bf smooth}: $u(t)$ and $u(t-1)$ should be close to each other for most of the time while allowing for occasional big shift. 
%
We propose finding temporal word embeddings as the solution of the following joint optimization problem:

\vspace{-0.3cm}
\small
\begin{align}
\label{eq:simfac}
\underset{U(1), \ldots, U(T)}{\text{min}} & \frac12 \sum_{t = 1}^T \| Y(t) - U(t)U(t)^T \|^2_F\\
\notag
&
+ \frac{\lambda}{2} \sum_{t=1}^T \| U(t) \|^2_F  + \frac{\tau}{2} \sum_{t = 2}^T \| U(t-1) - U(t) \|^2_F,
\end{align}
\normalsize
where $Y(t) = \text{PPMI}(t,L)$ and 
  $\lambda, \tau > 0$.
Here the penalty term $\| U(t) \|^2_F$ enforces the low-rank data-fidelity as widely adopted in previous literature. The key smoothing term $ \| U(t-1) - U(t) \|^2_F$ encourages the  word embeddings to be aligned. 
% he data fitting term in \eqref{eq:simfac}, much like in \eqref{eq:indfac}, forces the embeddings to change over time, based on the values in $Y(t)$. In addition, 
The parameter $\tau$ controls how fast we allow the embeddings to change; $\tau = 0$ enforces no alignment, and picking $\tau \to \infty$ converges to a static embedding with $U(1)=U(2) = \hdots = U(T)$. 
%A very large $\tau$ will force the embedding vector across time to be almost identical, diminishing the potential difference in the word-associations for different time points. 
%%%%%%%%%%%%%%%%%%%%%%%%%%%%%%%%
%%%%%%%%%%%%%%%%%%%%%%%%%%%%%%%%
%%%%%%%%%%%%%%%%%%%%%%%%%%%%%%%%
%%%%%%%%%%%%%%%%%%%%%%%%%%%%%%%%
%%%%%%%%%%%%%%%%%%%%%%%%%%%%%%%%
%%%%%%%%%%%%%%%%%%%%%%%%%%%%%%%%
%%%%%%%%%%%%%%%%%%%%%%%%%%%%%%%%
%%%%%%%%%%%%%%%%%%%%%%%%%%%%%%%%
%%%%%%%%%%%%%%%%%%%%%%%%%%%%%%%%
%%%%%%%%%%%%%%%%%%%%%%%%%%%%%%%%
%
%As a side product, the problem \eqref{eq:simfac} also addresses the alignment problem by introducing the smoothing regularization. 
Note that the methods of \cite{kulkarni2015statistically,hamilton2016diachronic} can be viewed as suboptimal solutions of \eqref{eq:simfac}, in that they optimize for each term separately.
For one, while the strategies in \cite{kulkarni2015statistically} and \cite{hamilton2016diachronic}  enforce alignment pairwise, we enforce alignment across \emph{all} time slices; that is, 
the final aligned solution $U(t)$ is influenced by not only $U(t-1)$ and $U(t+1)$, but every other embedding as well. This avoids the propagation of alignment errors caused by a specific time frame's subsampling.
Additionally, consider an extreme case in which word $w$ is absent from $\mathcal D_t$ but has similar meaning in both $t-1$ and $t+1$. Directly applying any matrix factorization technique to each time point would enforce $u_w(t)\approx \mathbf{0}$.
However, for the right choice of $\tau$, the solution $u_w(t)$ to \eqref{eq:simfac}    will be close to $u_w(t-1)$ and $u_w(t+1)$. 
%In this sense, \cite{kulkarni2015statistically,hamilton2016diachronic}  can be viewed as locally approximating our solution.
Overall, our method achieves high fidelity embeddings with a much smaller corpus, and in particular, in Section \ref{sec:expt}, we demonstrate that our embeddings  are robust against sudden undersampling of specific time slices.

\begin{figure*}[t]
        \begin{subfigure}[b]{0.24\textwidth}
                \centering
                {\includegraphics[width=1\textwidth]{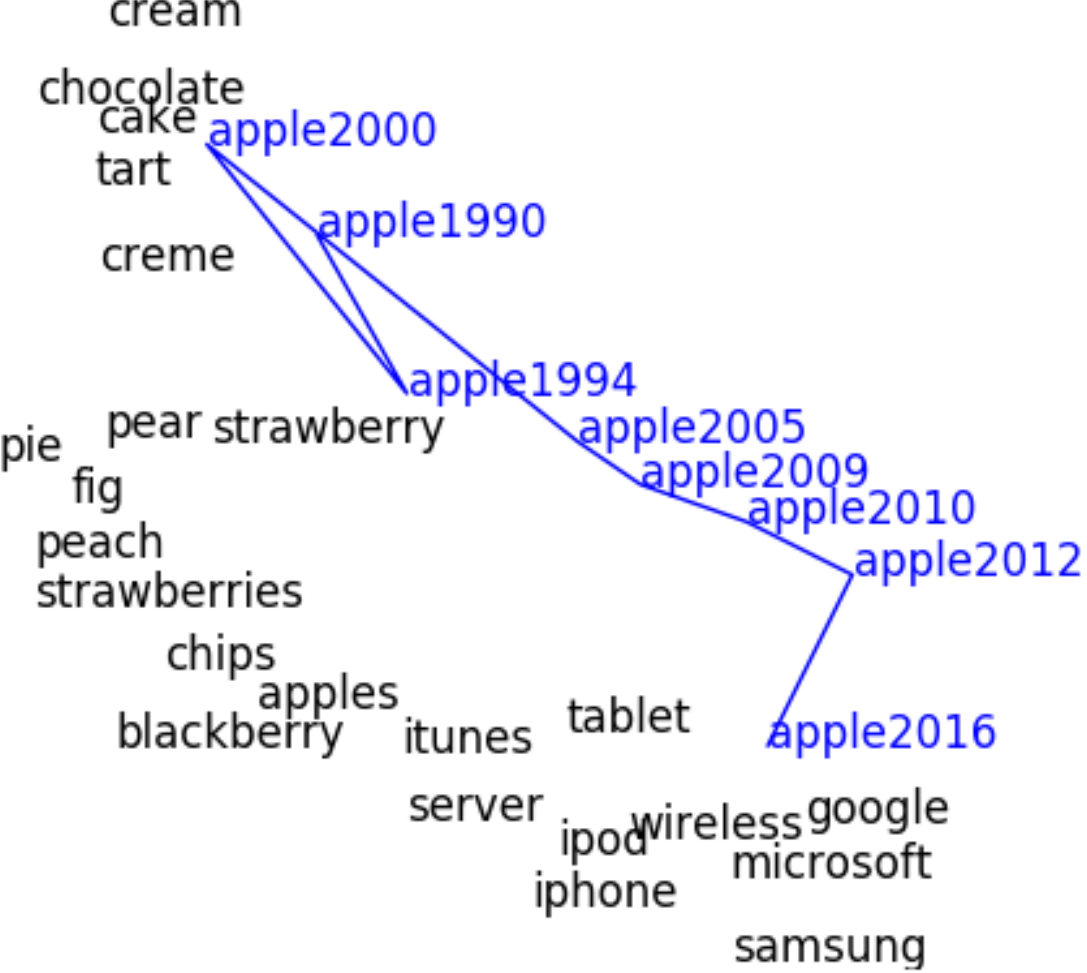}}
                \caption{apple}
                \label{fig:feature:a}
        \end{subfigure}%
        \hfill
        \begin{subfigure}[b]{0.24\textwidth}
                \centering
                {\includegraphics[width=1\textwidth]{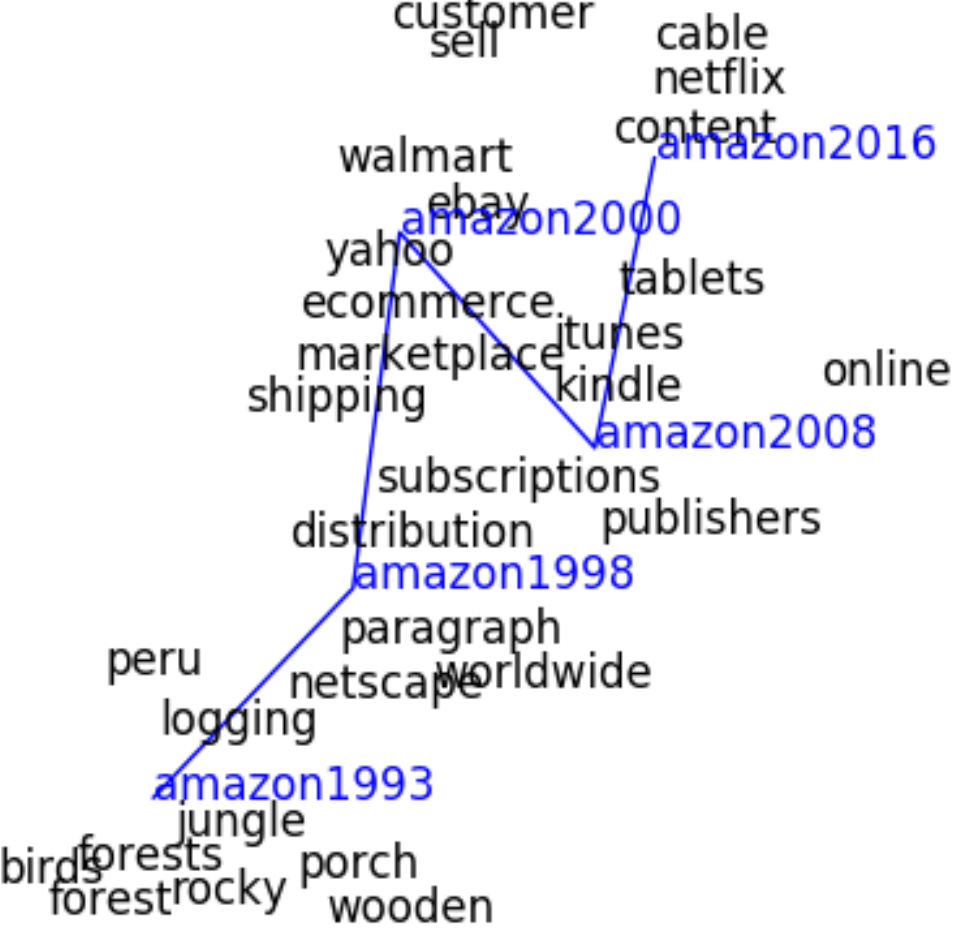}}
                \caption{amazon}
                \label{fig:feature:b}
        \end{subfigure}%
        \hfill
        \begin{subfigure}[b]{0.24\textwidth}
                \centering
                {\includegraphics[width=1\textwidth]{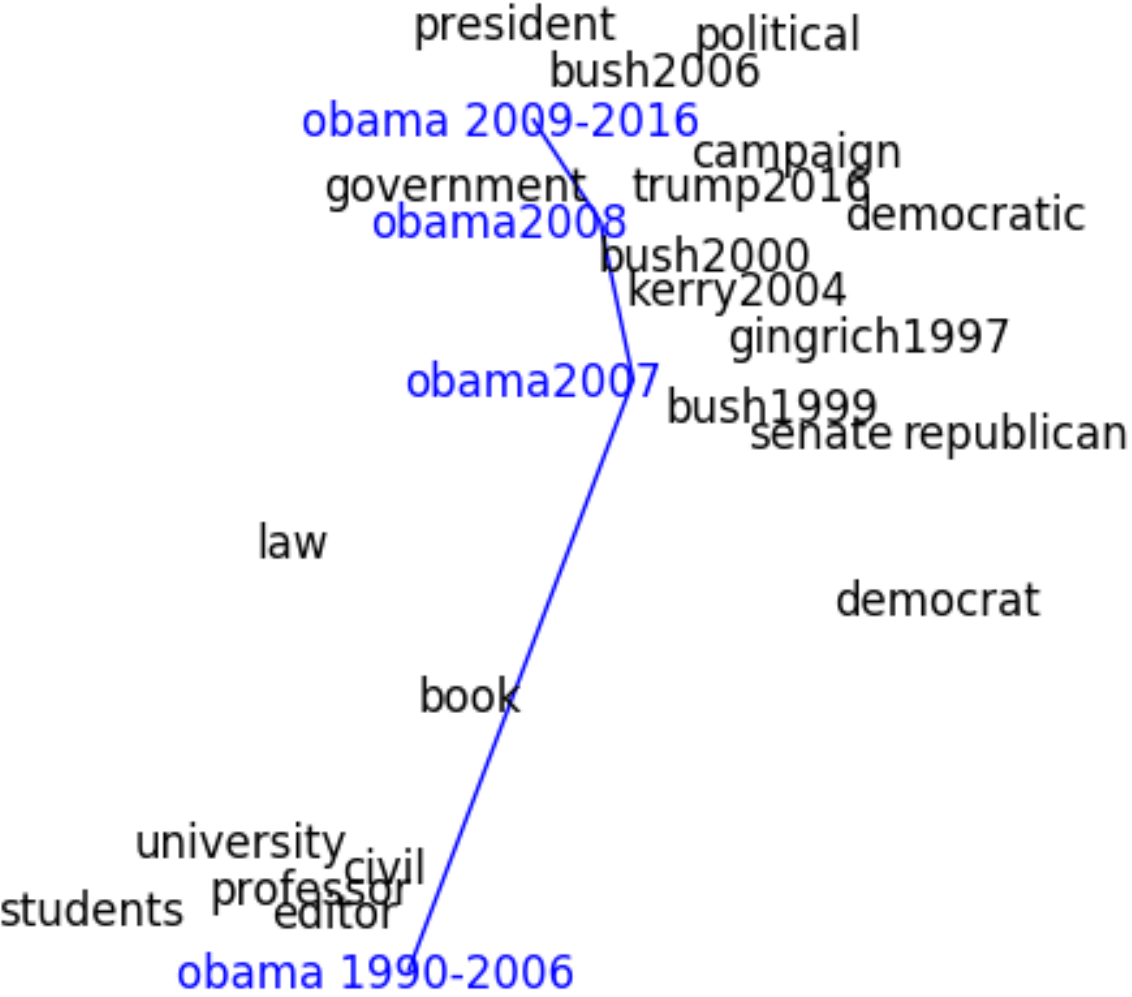}}
                \caption{obama}
                \label{fig:feature:c}
        \end{subfigure}%
        \hfill
        \begin{subfigure}[b]{0.24\textwidth}
                \centering
                {\includegraphics[width=1\textwidth]{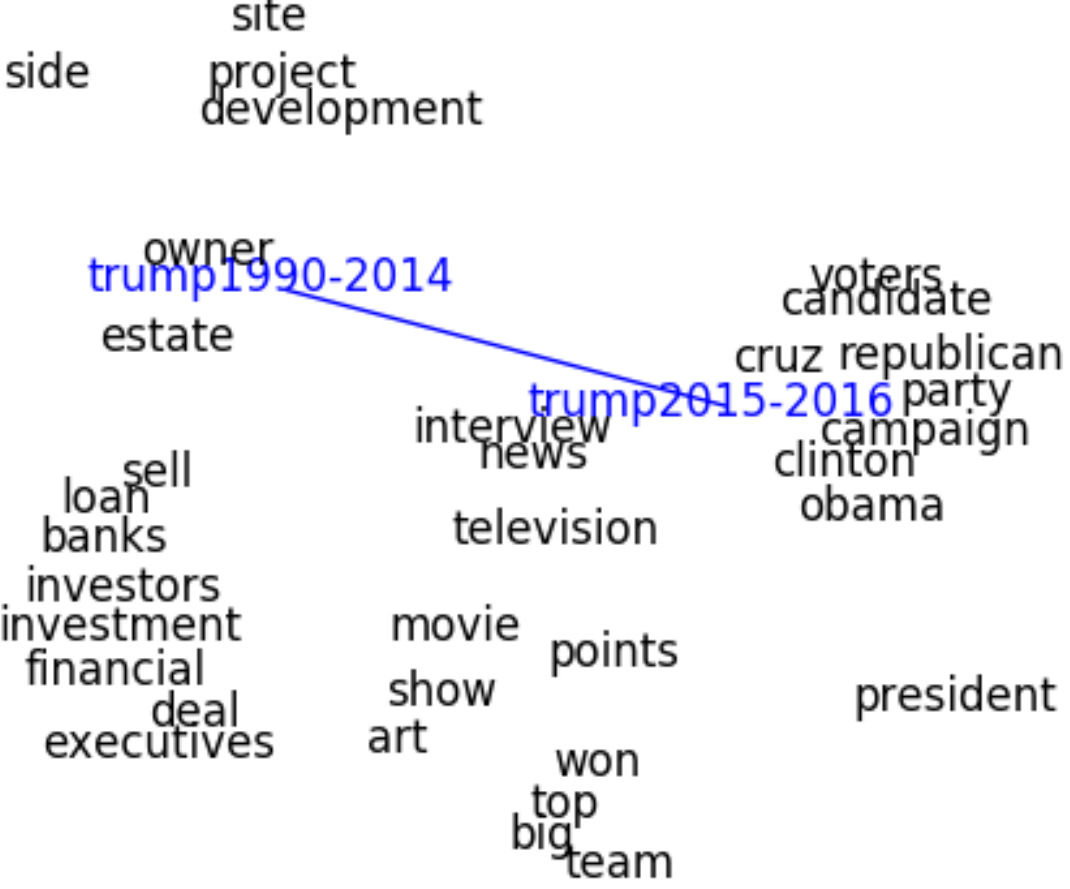}}
                \caption{trump}
                \label{fig:feature:d}
        \end{subfigure}%
%        \vspace{-0.2cm}
        \caption{Trajectories of brand names and people  through time: \texttt{apple}, \texttt{amazon}, \texttt{obama}, and \texttt{trump}.}
        \label{f-trajs}
%        \vspace{-0.1cm}
\end{figure*}
%\vspace{-0.3cm}
\section{Optimization}
\label{sec:bcd}
A key challenge in solving  \eqref{eq:simfac} is that for large $V$ and $T$, one might not be able to fit all the PPMI matrices $Y(1),\hdots, Y(T)$ in memory, even though $Y(t)$ is sparse. Therefore, for scalability, an obvious solution is to first decompose the objective across time, using alternating minimization to solve for $U(t)$ at each step:
%. An obvious choice is to use alternating minimization for each $t$, using gradient descent for each variable. First, without loss of generality consider solving for $U(t)$, with $t \neq 1 \mbox{ or } T$. The optimization problem to solve is then:

\vspace{-0.2cm}
\small
\begin{align}
\label{eq:singleU}
\min_{U(t)} &  \overbrace{\frac12  \| Y(t) - U(t)U(t)^T \|^2_F}^{f(U(t))} + \frac{\lambda}{2} \| U(t) \|^2_F \\
\notag
& + \frac{\tau}{2} \left( \| U(t-1) - U(t) \|^2_F + \| U(t) - U(t+1) \|^2_F \right)
\end{align}
\normalsize
for a specific $t$.
Note that $f(U(t))$ is quartic in $U(t)$, and thus even if we only solve for a fixed $t$,  \eqref{eq:singleU}  cannot be minimized analytically.
Thus, our only option is to solve  \eqref{eq:singleU}  iteratively 
 using a fast first-order method  such as gradient descent. The gradient of the first term alone is given by
\begin{equation}
\label{eq:gradut}
\nabla f(U(t)) =  -2Y(t)U(t) + 2U(t)U(t)^TU(t).
\end{equation}
Each gradient computation is of the order $O(nnz(Y(t))d + d^2V)$ (which is then nested in iteratively optimizing $U(t)$ for each $t$).\footnote{$nnz(\cdot)$ is the number of nonzeros in the matrix.}
In practical applications, $V$ is in the order of ten-thousands to hundred-thousands, and $T$ is in the order of tens to hundreds. 

Let us instead look at a slightly relaxed problem of minimizing 

\vspace{-0.3cm}
\small
\begin{align}
\label{eq:finfac}
\min_{U(t),W(t)  } & \frac12 \sum_{t = 1}^T \| Y(t) - U(t)W(t)^T \|^2_F + \frac{\gamma}{2} \sum_{t=1}^T \| U(t) - W(t) \|^2_F\\
\notag
&+ \frac{\lambda}{2} \sum_{t=1}^T \| U(t) \|^2_F + \frac{\tau}{2} \sum_{t = 2}^T \| U(t-1) - U(t) \|^2_F \\
\notag
&+ \frac{\lambda}{2} \sum_{t=1}^T \| W(t) \|^2_F + \frac{\tau}{2} \sum_{t = 2}^T \| W(t-1) - W(t) \|^2_F,
\end{align}
\normalsize
where variables $W(t), t=1,\hdots,T$ are introduced to break the symmetry of factorizing $Y(t)$. Now, minimizing for each $U(t)$ (and equivalently $W(t)$) is just the solution of a ridge regression problem, and can be solved in one step by setting the gradient of the objective of \eqref{eq:finfac} to 0, \ie~
$U(t)A = B$ where
\[
A = W(t)^TW(t) + (\gamma+\lambda+2\tau) I,
\]
\[
 B = Y(t)W(t)+\gamma W(t) {+}\tau (U(t-1) + U(t+1))
\]
for $t = 2,\hdots, T-1$, and constants adjusted for $t=0,T$.
Forming $A$ and $B$ requires $O(Vd^2 + nnz(Y(t))d)$, and solving $U(t)A=B$ can be done in $O(d^3)$ computations, in one step.
This can be further decomposed to row-by-row blocks of size $b$, by minimizing over a row block of $U(t)$ at a time, which reduces the complexity of forming $A$ and $B$ to $O(bd^2 + nnz(Y(t)[:b,:])d)$, \ie independent of $V$. This allows scaling for very large $V$, as only one row block of $Y(t)$ must be loaded at a time.

\paragraph{Block coordinate descent vs. stochastic gradient descent:}
The method described here is commonly referred to as \emph{block coordinate descent (BCD)} because it minimizes with respect to a single block ($U(t)$ or $W(t)$) at a time, and the block size can be made even smaller (a few rows of $U(t)$ or $W(t)$) to maintain scalability. The main appeal of BCD is scalability~\cite{yu2012scalable}; however, a main drawback is lack of convergence guarantees, even in the case of convex optimization~\cite{powell1973search}. In practice, however, BCD is highly successful and has been used in many applications \cite{wright2015coordinate}.
% (see \cite{wright2015coordinate} for examples).
Another choice of optimization is stochastic gradient descent (SGD), which decomposes the \emph{objective} as a sum of smaller terms. For example, the first term of \eqref{eq:finfac} can be written as a sum of terms, each using only one row of $Y(t)$:

\vspace{-0.3cm}
\begin{equation}
f(U(t)) = \sum_{i=1}^V \|Y(t)[i,:] - u_i(t)^TW(t)\|_F^2.
\end{equation}

The complexity at first glance is smaller than that of BCD; however, SGD comes with the well-documented issues of slow progress and hard-to-tune step sizes, and in practice, can be much slower for matrix factorization applications \cite{rao2015collaborative, yu2012scalable}.
However, we point out that the choice of the optimization method is agnostic to our model; anything that successfully solves \eqref{eq:simfac} should lead to an equally successful embedding.

\section{Experimental Dataset and Setup}\label{sec:eval}
%\subsection{Details on computing embeddings}
In this section we describe the specific procedure used to generate embeddings for the next two sections. 

\noindent\textbf{News article dataset}:~
First, we crawl a total of 99,872 articles from the New York Times, published between January 1990 and July 2016.\footnote{The data is available at: https://sites.google.com/site/zijunyaorutgers/.}
In addition to the text, we also collected metadata including title, author, release date, and section label (\eg, Business, Sports, Technology); in total, there are 59 such sections.
We use yearly time slices, dividing the corpus into $T = 27$ partitions.
After removing rare words (fewer than 200 occurrences in all articles across time) and stop words, our vocabulary consists of $V = 20,936$ unique words. We then compute a co-occurrence matrix for each time slice $t$ with a window size $L = 5$, which is then used to compute the PPMI matrix as outlined in \eqref{ppmidefinition}. All the embedding methods that we compared against are  trained on this same dataset.

\noindent\textbf{Training details for our algorithm}:~  We perform a grid search to find the best regularization and optimization parameters. As a result of our search, we obtain  $\lambda = 10$, $\tau = \gamma = 50$, and run for 5 epochs (5 complete pass over all time slices, and all rows and columns of $Y(t)$). Interestingly, setting $\lambda = 0$ also yielded good results, but required more iterations to converge. The block variable is one matrix ($U(t)$ or $V(t)$ for a specific $t$).

\noindent{\bf Distance metric}:~
All distances between two words are calculated by the cosine similarity between embedding vectors:
\vspace{-0.2cm}
\begin{equation}
\label{cosineSim}
\text{similarity}(a,b) =\text{cosine}(u_{a},u_{b})=\frac{u_{a}^Tu_{b}}{\|u_{a}\|_2\cdot\|u_{b}\|_2},
\end{equation}
\vspace{-0.2cm}
where $u_{a}$ and $u_{b}$ are the embeddings of words $a$ and $b$.

\section{Qualitative evaluation}\label{sec:qualitative}
The embeddings we learn reveal interesting patterns in the shift of word semantics, cross-time semantic analogy, and popularity trends of concepts from the news corpus.
%We first demonstrate some of these findings in this section.
%
\subsection{Trajectory visualization}
The trajectory of a word in the (properly aligned) embedded space provides tools to understand the shift in meanings of words over time.  
%Studying word trajectories helps us understand the shift in meanings of words over time. 
This can help broader applications, such as capturing and quantifying linguistic evolution, characterizing brands and people, and analyzing emerging association between certain words. 

%As new concepts and information appear over time, words are dominated by different meanings.
%Here we provide ways of visualizing trajectories, first by plotting a 2-D projection and second by listing evolving neighbors.
Figure \ref{f-trajs} shows the trajectories of a set of example words. We plot the 2-D t-SNE projection of each word's temporal embedding across time. We also plot the closest words to the target word from each time slice. We pick four words of interest: \texttt{apple} and \texttt{amazon} as emerging corporate names while originally referring to a fruit and a rainforest, and \texttt{obama} and \texttt{trump} as people with changing professional roles.

%Figure \ref{f-trajs} shows the first visualization strategy which forms the trajectory of each interest word $w$ by a series of its temporal embedding through time $u_w(1), \hdots, u_w(T)$ and traverses every year for close words.
%Four interest words (\texttt{apple} and \texttt{amazon} for brand, \texttt{obama} and \texttt{trump} for person) are picked.
%Meanwhile, we provide their temporal closest words (randomly sampled for visualizing purpose) in different time state for illustrating their temporal meaning.
%We plot 2-D t-SNE projections of $d$-dimensional temporal embedding to enable the visualization.
In all cases, the embeddings illustrate significant semantic shifts of the words of interest during this 27-year time frame. 
%Also each interest word is surrounded by its temporal closest words.
We see \texttt{apple} shift from a fruit and dessert ingredient to space of technology. Interestingly, there is a spike in 1994 in the trajectory, when Apple led a short tide of discussion because of the replacement of the CEO and a collaboration with IBM; then the association shifted back to neighborhood of fruit and dessert until the recovery by Steve Jobs in early 2000s. Similarly, \texttt{amazon} shifts from a forest to an e-commerce company, finally landing in 2016 as a content creation and online-streaming provider due to the popularity of its  Prime Video service. The US president names, \texttt{obama} and \texttt{trump}, are most telling, shifting from their pre-presidential lives (Obama as a civil rights attorney and university professor; Trump as a real estate developer and TV celebrity) to the political sphere.
Overall, Figure \ref{f-trajs} demonstrates that first, our temporal word embeddings can well capture the semantic shifts of words across time, and second,
our model provides high alignment quality in that same-meaning words across different years have geometrically close embeddings, without having to solve a separate optimization problem for alignment. 
%\vspace{-0.2cm}

\subsection{Equivalence searching}
Another key advantage of word alignment is the ability to find conceptually ``equivalent'' items or people over time. We provide examples in the field of technology, official roles, and sports professionals.
In this type of test, we create a query consisting of a word-year pair that is particularly the representative of that word in that year, and look for other word-year pairs in its vicinity, for different years.

%In Table \ref{t-trajectory} we fix a query word and a year, chosen to be a particularly representative emerging technological item of that time, and find  words across time that are close to the embedding of that query word and year. 
Table \ref{t-trajectory} lists the closest words ($top$-1) of each year to the query vector.
%temporal vector of technological items over the time periods, 
For visualization purpose we lump semantically similar words together.
For example, the first column shows that \texttt{iphone} in 2012 is closely associated with smartphones in recent years, but is close to words such as \texttt{desktop} and \texttt{macintosh} in the 90's; interestingly, \texttt{telephone} never appears, suggesting the iPhone serves people more as a portable computer than a calling device.
%We use the popularization of technologies: \texttt{iphone}, \texttt{twitter}, and \texttt{mp3}.
%For \texttt{iphone}, though PCs were available in the early 90s, their domination seemed to occur closer to late 1990s, and are now mostly replaced by smartphones.
As another example, by looking at the trajectory of \texttt{twitter}, we see the evolution of news sources, from  TV \& radio news broadcasts in the 90s to chatrooms, websites, and emails in the early 2000s, blogs in the late 2000s, and finally tweets today. 
%Again, interestingly, \texttt{letters} or \texttt{telegrams} never appear, suggesting twitter is more of a news source than a communication device.
%we can observe that as an emerging tool for people to receive news, communicate with each other, and share thoughts, it is replacing 
%
The last example is fairly obvious; \texttt{mp3} represents the main form of which music is consumed in 2000, replacing disk and stereo in 1990s  ( \texttt{cassette} also appears in  $top$-3) and is later replaced by  online streaming. We can see a one-year spike of Napster which was shut down  because of copyright infringement\footnote{Napster ended its streaming service in 2001, so our equivalence is captured 2 years late; this delay could be because though the event happened in 2001, the legal ramifications were analyzed heavily in subsequent years.}, and later a new streaming service - iTunes.

\begin{table}
\caption{Equivalent technologies through time: \texttt{iphone}, \texttt{twitter}, and \texttt{mp3}.}
{\small
\begin{tabular}{|c|l|l|l|}
\hline
Query& \texttt{iphone}, 2012 & \texttt{twitter}, 2012 & \texttt{mp3}, 2000 \\\hline
90-94	& \multirow{3}{.11\textwidth}
{
\begin{minipage}[t]{.1\textwidth}\raggedright desktop, pc, dos, macintosh, software \end{minipage}
}
& 
\multirow{2}{.11\textwidth}
{
\begin{minipage}
[t]{.1\textwidth}\raggedright broadcast, cnn, bulletin, tv, radio, messages, correspondents\end{minipage}
}& 
\begin{minipage}[t]{.1\textwidth}\raggedright stereo, disk, disks, audio\end{minipage}
\\[.5cm]\cline{1-1}\cline{4-4}
95-96 &&&
\multirow{3}{.11\textwidth}{mp3}\\[.5cm]\cline{1-1}\cline{3-3}
97 &&
\multirow{4}{.11\textwidth}
{
\begin{minipage}{.1\textwidth}\raggedright chat, messages, emails, web\end{minipage}
} 
& \\\cline{1-2}
98-02&
\multirow{3}{.11\textwidth}{pc} 
&&\\\cline{1-1}\cline{4-4}
03  &&&napster\\\cline{1-1}\cline{4-4}
04&&&mp3\\\cline{1-4}
05-06 &ipod&
\multirow{2}{.11\textwidth}
{
\begin{minipage}{.1\textwidth}\raggedright blog, posted\end{minipage}
} 
&\multirow{3}{.11\textwidth}
{
\begin{minipage}{.1\textwidth}\raggedright itunes, downloaded\end{minipage}
} 
\\\cline{1-2}
07-08&
\multirow{2}{.11\textwidth}{iphone} 
&& \\\cline{1-1}\cline{3-3}
09-12&&
\multirow{2}{.11\textwidth}
{
\begin{minipage}{.1\textwidth}\raggedright twitter \end{minipage}
} 
&\\\cline{1-2}
13-16&
\begin{minipage}[t]{.1\textwidth}\raggedright smartphone, iphone\end{minipage}
&&\\[.4cm]\hline
\end{tabular}
}
\label{t-trajectory}
\vspace{-0.3cm}
\end{table}

\begin{table}
\caption{``Who governed?'' The closest word to \texttt{obama} at year 2016 (role as president of United State) and \texttt{blasio} at year 2015 (role as mayor of New York City (NYC)). The stars indicate incorrect answers.}
\small
\begin{tabular}{|c|l|l|}
\hline
Question 	& US president & NYC mayor\\\hline
Query		& \texttt{obama}, 2016& \texttt{blasio}, 2015\\\hline
90-92		&bush												&	\multirow{2}{.11\textwidth}{dinkins}\\\cline{1-2}
93			&\multirow{2}{.11\textwidth}{clinton}	&														\\\cline{1-1}\cline{3-3}
94-00		&														&\multirow{2}{.11\textwidth}{giuliani}\\\cline{1-2}
01			&\multirow{5}{.11\textwidth}{bush}		&														\\\cline{1-1}\cline{3-3}
02-05		&														&bloomberg										\\\cline{1-1}\cline{3-3}
06			&														& n/a*												\\\cline{1-1}\cline{3-3}
07			&														&	\multirow{3}{.11\textwidth}{bloomberg}\\\cline{1-1}
08			&														&														\\\cline{1-2}
09-10		&\multirow{4}{.11\textwidth}{obama}	&														\\\cline{1-1}\cline{3-3}
11			&														&cuomo*											\\\cline{1-1}\cline{3-3}
12			&														&bloomberg										\\\cline{1-1}\cline{3-3}
13-16		&														&blasio												\\
\hline
\end{tabular}
\normalsize
\label{t-president-mayor}
\vspace{-0.3cm}
\end{table}

Next, we use embeddings to identify people in political roles.
%search the people of same roles in different time periods, such as a political role (e.g., US president and NYC mayor), and sports players (e.g., rank 1 tennis player).
Table \ref{t-president-mayor} attempts to discover who is the U.S. president\footnote{All data was scraped about half a year before Donald Trump was elected as U.S. president in 2016.} and New York City mayor\footnote{We intentionally choose New York City because it is the most heavily discussed city in the New York Times.} of the time, using as query \texttt{obama} in 2016 and \texttt{blasio} in 2015.
For president, only the closest word from each year is listed, and is always correct (accounting for the election years).
For mayor, the top-1 closet word is shown unless it is \texttt{mayor}, in which case the second word is shown. 
We can see that both the roles of US president and NYC mayor have been well searched for different persons in their terms of service. We see that the embedding for the President is consistent, and for the most part, so is that of the mayor of NYC. 
%\red{For finding the  NYC mayors, although we do not find the correct answer in the first relevant word for 2011 and 2006, the answers are still be found in $top$-5 results.} 
In 2011, \texttt{cuomo} is also partially relevant since he was the governor of NY state.
We did not find any relevant words in query NYC mayor in year 2006.
% role of people in different time periods, in politics (table \ref{t-president-mayor}) and sports (table \ref{t-tennis}). 
%Table \ref{t-president-mayor} attempts to discover who is the U.S. president and New York City mayor of that time.
%For all but two, the correct NYC mayor is extracted; two wrong answers were found are starred.
%\todo{should we compare against stanford's result, so that it's more impressive?}
%In instances when the object has a title (president, mayor), we find this test to be extremely robust. See also table {xx}.

Finally, we search for equivalences in sports, repeating the experiment for the ATP rank 1 male tennis player as shown in Table \ref{t-tennis}.
In the case of president and mayor, we are heavily assisted by the fact that they are commonly referred to by  a title: ``President Obama'' and ``Mayor de Blasio''. Tennis champions, on the other hand, are not referred by titles. Still, a surprising number of correct champions appear as the closest words, and all the names are those of famous tennis players for the given time period.
A more exhaustive empirical study of alignment quality is provided in Section \ref{sec:expt}.

\begin{table}
\caption{``Who was the ATP No.1 ranked male player?'' The closest word to \texttt{nadal} at year 2010 for each year is listed. The correct answer is based on ATP year-end ranking and are bolded in the table.}
\small
\begin{tabular}{|c|c|c|c|c|c|}
\hline
year & 1990 & 1991 & 1992& 1993  \\\hline
word &\textbf{edberg} & lendl  & sampras  &\textbf{ sampras}  \\\hline\hline
1994 & 1995 & 1996 & 1997& 1998  \\\hline
 \textbf{sampras} & \textbf{sampras}  & ivanisevic  & \textbf{sampras} & \textbf{sampras} \\\hline\hline
1999 & 2000 & 2001 & 2002& 2003  \\\hline
\textbf{sampras} & sampras & agassi  &  capriati & \textbf{roddick}  \\\hline\hline
2004 & 2005 & 2006 & 2007& 2008  \\\hline
\textbf{federer} &\textbf{federer} & roddick  & \textbf{federer}  & \textbf{nadal}  \\\hline\hline
2009 & 2010 & 2011 & 2012 & 2013  \\\hline
\textbf{federer} & \textbf{nadal} & \textbf{djokovic} & federer  & federer  \\\hline\hline
2014& 2015\\\cline{1-2}
federer &\textbf{djokovic}\\\cline{1-2}
\end{tabular}
\normalsize
\label{t-tennis}
\vspace{-0.3cm}
\end{table}

\subsection{Popularity determination}
%We begin by looking at the word embedding norms, which can be used for changepoint detection. 
It has often been observed that word embeddings computed by factorizing PMI matrices have norms that grow with word frequency \cite{glove,arora2015rand}.
%As discussed in \cite{kulkarni2015statistically}, these word vector norms can be viewed as a distributional time series that can detect change points (sudden semantic shifts or emergences), often more robustly than word frequency.
These word vector norms across time can be viewed as a time series for detecting the trending concepts (\eg, sudden semantic shifts or emergences) behind words, with more robustness than word frequency.

\begin{figure}
\begin{center}
\includegraphics[height=2.5cm]{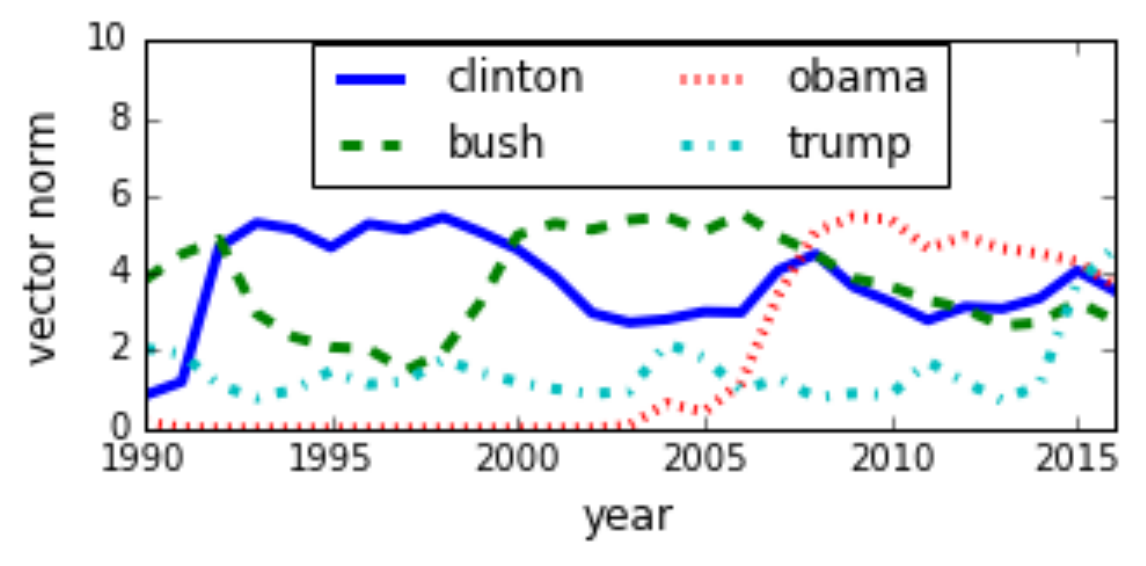}

\includegraphics[height=2.5cm]{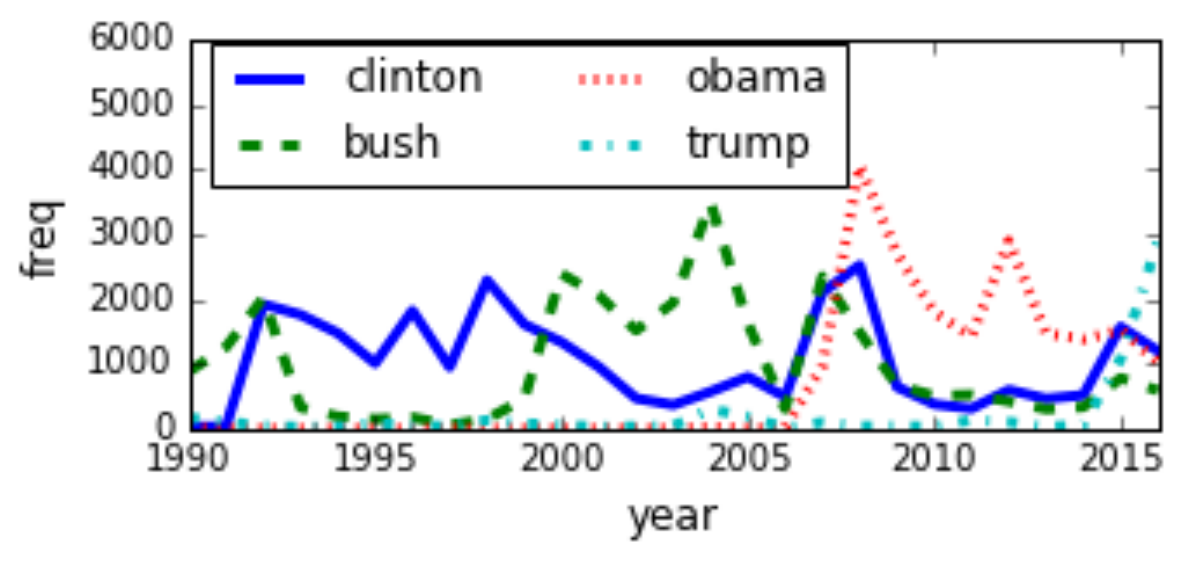}
\end{center}
\vspace{-3mm}
\caption{Norm (top) and relative frequency (bottom) throughout years 1990-2016. We select the names of U.S presidents within this time frame - \texttt{clinton}, \texttt{bush}, \texttt{obama}, and \texttt{trump}. We note that \texttt{bush} could refer to George H. W. Bush (1989-1993) or George W. Bush (2001-2009). Clinton could refer to Bill Clinton (1993-2001) or Hillary Clinton (U.S. secretary of state in 2009-2013 and U.S. presidential candidate in 2014-2016).}
\label{f-presidents}
\vspace{-0.3cm}
\end{figure}

\begin{figure}
\begin{center}
\includegraphics[height=2.5cm]{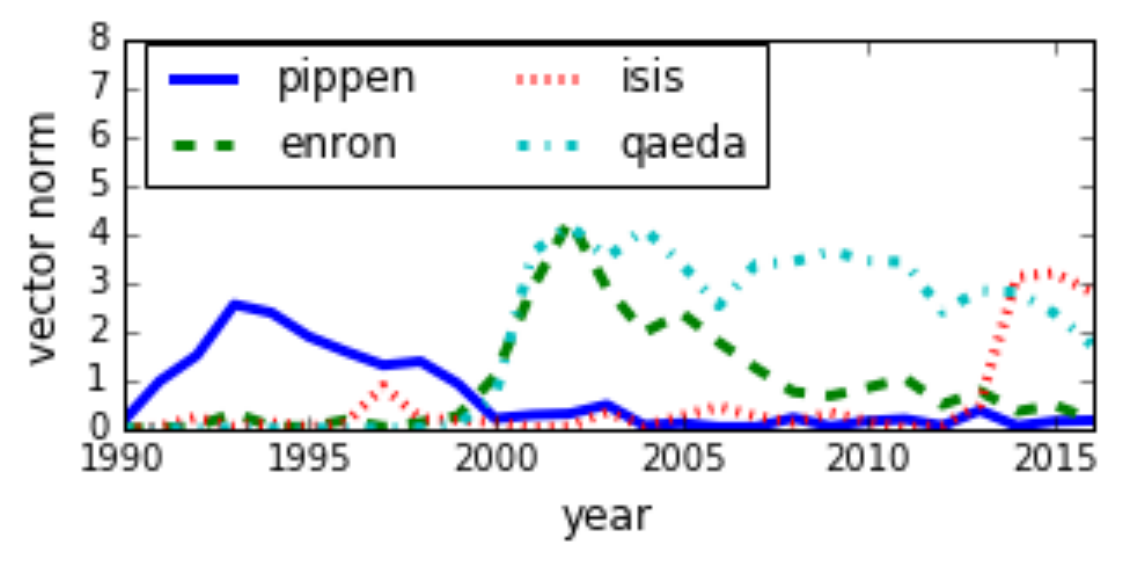}

\includegraphics[height=2.5cm]{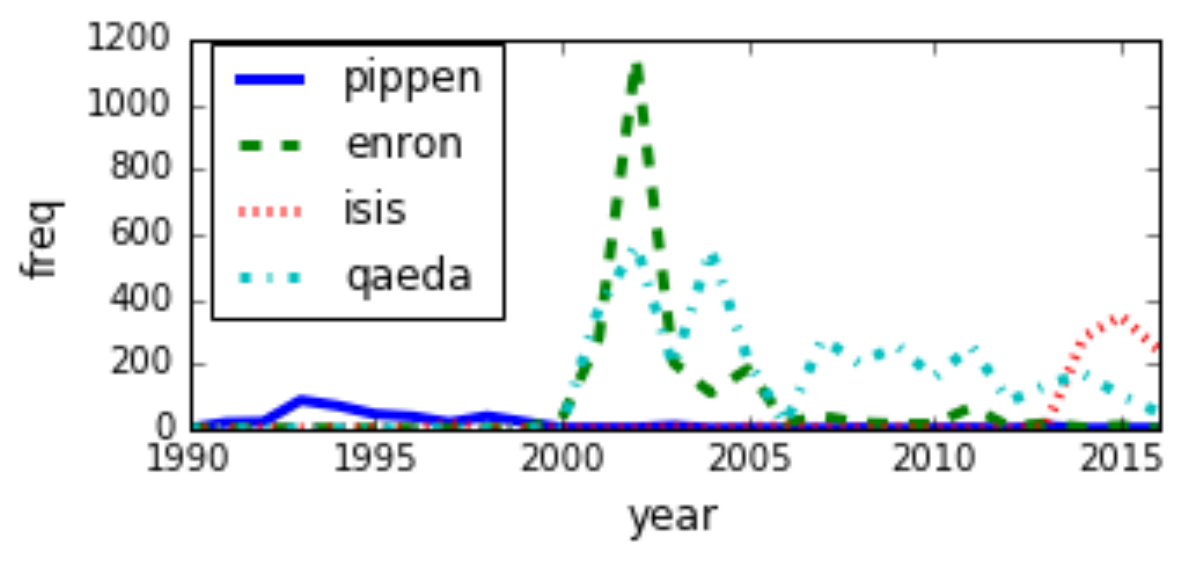}
\end{center}
\vspace{-3mm}
\caption{Norm (top) and relative frequency (bottom) per year in corpus of major event keywords: \texttt{pippen} for basketball stardom, \texttt{enron} for corporation  scandal, \texttt{qaeda} and \texttt{isis} for emerging terrorism groups.}
\label{f-events}
%\vspace{-0.3cm}
\end{figure}

Figures \ref{f-presidents} and \ref{f-events} illustrate the comparison between embedding norm and frequency for determining \emph{concept} popularity per year, determined by key words in the New York Times corpus.
Generally, comparing to frequencies which are much more sporadic and noisy, we note that the norm of our embeddings encourages smoothness and normalization while being indicative of the periods when the corresponding words were making news rounds.
In Figure \ref{f-presidents}, the embedding norms display nearly even 4-year humps corresponding to each president's term. 
In every term, the name of each current president becomes a trending concept which plays an important role in the information structure at the time. Two interesting observations can be gleaned.
First, since Hillary Clinton continuously served as Secretary of State during 2009-2013, the popularity of \texttt{clinton} was preserved; however it was still not as popular as president \texttt{obama}.
Second, because of the presidential campaign, \texttt{trump} in 2016 has a rising popularity that greatly surpasses that of his former role as a  business man, and eventually surpasses his opponent \texttt{clinton} in terms of news coverage.

In Figure \ref{f-events}, we can see smooth rises and falls of temporary phenomena (the \texttt{enron} scandal and \texttt{qaeda} rises). For \texttt{qaeda}, we see that there is a jump in 2001, and then it remains steady with a small decline.
In contrast, \texttt{enron} shows a sharper decline, 
as despite its temporal hype, it did not linger in the news long.
Note also the stability of using norms to track popularity over frequency, which spikes for \texttt{enron} above \texttt{qaeda}, although 9/11 was far more news-prevalent than the corporation's scandalous decline.
For the basketball star \texttt{pippen}, although his publicity (\eg, frequency) was relatively fewer than business terms, his popularity is still recognized by the enhancement in vector norm.
For another term \texttt{isis}, we can see that it begins to replace \texttt{qaeda} as the ``trending terrorist organization''
%\footnote{for want of a better description}
in news media.
%and a replace like behavior for enduring concepts (like terrorist cells). 
%Note, however, that the corresponding frequency plots, while showing peaks at the right points, are much more sporadic, and thus noisy.

%\vspace{-0.2cm}

\section{Quantitative evaluation} \label{sec:expt}

In this section, we empirically evaluate our proposed Dynamic Word2Vec model (DW2V) against other temporal word embedding methods.\footnote{The testsets are available at: https://sites.google.com/site/zijunyaorutgers/.} In all cases, we set the embedding dimension to $d = 50$.
We have the following baselines:
%
%\noindent{\bf Baseline Algorithms}:~
%We compared our method (DW2V) against various baseline static and temporal word embedding approaches:
%
%For evaluating the accuracy of word embedding and consistency of different time slots, we compare our Dynamic Word2Vec (DW2V) method with baselines including classic Skipgram method and popular time-aware word embedding methods.
%
\begin{itemize} [leftmargin=*]
\item \textbf{Static-Word2Vec (SW2V)}: the standard word2vec embeddings \cite{mikolov2013distributed}, trained on the entire corpus and ignoring time information. 
%The embedding dimension is $d=50$. 
%
%\item \textbf{Separated-Word2Vec (Sep-W2V)}: the corpus in each time slots (e.g., year 1990) is trained independently, and no alignment method is applied.
%
\item \textbf{Transformed-Word2Vec (TW2V) \cite{kulkarni2015statistically}}: the embeddings $U(t)$ are first trained separately by factorizing PPMI matrix for each year $t$, and then transformed by optimizing a linear transformation matrix which minimizes the distance between $u_w(t)$ and $u_w(t')$ for the $k=30$ nearest words' embeddings to the querying word $w$. 
%The embedding dimension is $d=50$ and $k=10$. 
%
%based on the separated training on the corpus of each time slot, an alignment method for each word is applied by calculating a linear transformation matrix which minimizes its $k$ nearest words' embeddings differences between original and destination time slots.
\item \textbf{Aligned-Word2Vec (AW2V) \cite{hamilton2016diachronic}}:  the embeddings $U(t)$ are first trained by factorizing the PPMI matrix for each year $t$, and then aligned by searching for the best othornormal transformation between $U(t)$ and $U(t+1)$. 
%We set the embedding dimension to be $d=50$. 
% a transformation matrix is computed by solving Orthogonal Procrustes problem for aligning the word embeddings between adjacent time slots.  
%\item \textbf{Dynamic-Word2Vec (Dynamic-W2V-2)} Our method, which solves \eqref{eq:finfac}. 
\end{itemize}

\subsection{Semantic similarity}
One of the most important properties of a word embedding is how accurately it carries the meaning of words.
Therefore, we develop a test to see if words can be categorized by meaning based on embeddings.
The news articles we collected are tagged with their ``sections'' such as \textit{Business}, \textit{Sports}. This information can be used to determine temporal word meanings.
% In analyzing news media, one method to gather temporal word meanings is to find its yearly frequency of usage in article sections (\eg, \textit{Business}, \textit{Sports}). 
It is important to note that this information is not used in the word embedding learning.
For example, % \texttt{nasdaq} occurs 97\% of the time in \textit{Business}  in 1991, we  say that \texttt{nasdaq} is strongly associated with  business  in that year.
we see that \texttt{amazon} occurs 41\% of the time in \textit{World} in 1995, associating strongly with the rainforest (not part of the USA), and 50\% of the time in \textit{Technology} in 2012, associating strongly with e-commerce.
%the semantic shift of \texttt{apple} can be detected by observing its relative frequency shift from food related sections to \textit{Business} and \textit{Technology}.
%For testing the result of clustering, we use the news article section (e.g., Arts, Sport, Technology) in which the word occurs to label words with themes (e.g., ``quarterback'' is a Sport related word).
We thus use this to establish a ground truth of word category, by identifying words in years that are exceptionally numerous in one particular news section. That is, if a word is extremely frequent in a particular section, we associate that word with that section and use that as ground truth. 
We select the 11 most popular and discriminative sections\footnote{Arts, Business, Fashion \& Style, Health, Home \& Garden, Real Estate, Science, Sports, Technology, U.S., World.} of the New York Times, and for each section $s$ 
and each word $w$ in year $t$, we compute its percentage $p$ of occurrences in each section.
%We say that  $w$ in year $t$ is associated with section $s$ with strength $p$.
To avoid duplicated word-time-section $<w,t,s>$ triplets, for a particular $w$ and $s$ we only keep the year of the largest strength, and additionally filter away any triplet with strength less than $p = 35\%$. Note that a random uniform distribution of words would result in it appearing about $9\%$ of the time in each section, and our threshold is about 4 times that quantity. We do this to say with sufficient confidence that such associations can be treated as ground truth.

\begin{table}%[h]
\abovecaptionskip
\belowcaptionskip
%\addtolength{\tabcolsep}{-3pt}
\caption{Normailized Mutual Information (NMI).}\label{table:NMI}
%\vspace{-0.3cm}
%\scriptsize
%\addtolength{\tabcolsep}{-5pt}
\begin{center}
\small
    \begin{tabular}{  c | c | c | c}
    \hline\hline
    Method & 10 Clusters & 15 Clusters & 20 Clusters \\ \hline
    SW2V  & 0.6736 & 0.6867 & 0.6713\\
    TW2V  & 0.5175 & 0.5221 & 0.5130\\
    AW2V  & 0.6580 & 0.6618 & 0.6386\\
   	DW2V & \textbf{0.7175} & \textbf{0.7162} & \textbf{0.6906}\\
    \hline\hline
    \end{tabular}
\normalsize
\end{center}
%\vspace{-0.3cm}
\end{table}

\begin{table}%[h]
\abovecaptionskip
\belowcaptionskip
%\addtolength{\tabcolsep}{-3pt}
\caption{F-measure ($F_\beta$).}\label{table:Fmeasure}
%\vspace{-0.3cm}
%\scriptsize
%\addtolength{\tabcolsep}{-5pt}
\begin{center}
\small
    \begin{tabular}{  c | c | c | c}
    \hline\hline
    Method & 10 Clusters & 15 Clusters & 20 Clusters \\ \hline
    SW2V  & 0.6163 & 0.7147 & 0.7214\\
    TW2V  & 0.4584 & 0.5072 & 0.5373\\
    AW2V  & 0.6530 & 0.7115 & 0.7187\\
   	DW2V & \textbf{0.6949} & \textbf{0.7515} & \textbf{0.7585}\\
    \hline\hline
    \end{tabular}
\normalsize    
\end{center}
\vspace{-0.2cm}
\end{table} 

% denoted as $<(w,\hat{t},s),p>$.
%To filter for strong associations, we select all triplets $(w,\hat{t},s)$ with strength $p>35\%$.
% For each word/section pair, we select the year of which that percentage is largest, forming $11 \times V$ triplets (word, year $t$, section). We then remove all triplets for which the percentage is under a threshold ($35\%$). 
%To further filter, for any section with more than 200 associated words in various times, we rank all words according to its maximum (over time) association with that section, and keep only  $(w,t)$ pairs where $w$  is within the first top 200 words (for any time).
To limit the size differences among categories, for every section $s$ with more than 200 qualified triplets, we keep the $top$-200 words by strength.
In total, this results in 1888 triplets across 11 sections, where every word-year pair is strongly associated with a section as its true category.
%, and indicates a strong section association.
% we count the maximum frequency of words appear in each section across all years.
%If a word appears heavily in a section (that is, 35\% or more of its occurrences appeared in that section for a year $t$) we  claim that this word in year $t$ is strongly related to this section, and give it that section label.
%Therefore, for every word in every section, we record the largest occurrence percentage as well as the time slots over the history, and label the word with a section if the percentage exceeds a minimum threshold (40\%).
%
%For example, if word ``apple" has 40\% occurrence in ``Technology" section in 2015 which is the largest percentage over all the years for 'Technology' section, then we use the embedding of ``apple" in 2015 to represent its technology related meaning, and denote it as ``apple-2015" which is labeled ``Technology".
%In this way, we obtain a set of word-year pairs which are labeled according to a strong section preference.

We then apply spherical k-means, which uses cosine similarity between embeddings as the distance function for clustering, with $K = $ 10, 15, and 20 clusters. We use two metrics to evaluate the clustering results:

%
%
%A good embedding learning would make same theme words have similar embedding values, therefore, the clustering can differentiate words by their themes.
%
%We evaluate the result of cluster with following metrics:
%
\begin{itemize} [leftmargin=*]
\item 
%Table \ref{table:NMI} gives the 
\emph{Normalized Mutual Information (NMI)}, defined as
\begin{equation}
\label{eq:NMI}
NMI(L,C)=\frac{I(L;C)}{[H(L)+H(C)]/2},
\end{equation}
where $L$ represents the set of labels and $C$  the set of clusters. $I(L;C)$ denotes the sum of mutual information between any cluster $c_i$ and any label $l_j$, and $H(L)$ and $H(C)$ the entropy for labels and clusters, respectively.
This metric evaluates the purity of clustering results from an information-theoretic perspective.

%\item \textbf{Rand Index (RI)}
%\begin{equation}
%\label{RI}
%RI=\frac{TP+TN}{N(N-1)/2},
%\end{equation}
%where $TP$ (true positive) denotes the number of word pairs with same labels are clustered into same clusters, $TN$ (true negative) denotes the number word pairs with different labels are clustered into different clusters. $N$ is the number of words.
\item
%Table \ref{Fmeasure} gives the 
\emph{F$_\beta$-measure ($\mathbf{F_\beta}$)}, defined as
\begin{equation}
\label{eq:Fmeasure}
F_\beta=\frac{(\beta^2+1)PR}{\beta^2P+R},
\end{equation}
where $P=\frac{TP}{TP+FP}$ denotes the precision and $R=\frac{TP}{TP+FN}$ denotes the recall.
(TP/FP = true/false positive, TN/FN = true/false negative.)
As an alternative method to evaluate clustering, we can view every pair of words as a series of decisions. Pick any two $(w,t)$ pairs. If they are clustered together and additionally have the same section label, this is a correct decision; otherwise, the clustering performed a wrong decision.
% If the clustering result agree with label consistency (true positive $TP$ and true negative $TN$), it makes a right decision, otherwise (false positive $FP$ and false negative $FN$), it makes a wrong decision.
The metric $F_\beta$ measures accuracy as the ($\beta$-weighted) harmonic mean of the precision and recall.
%Therefore, precision $P=\frac{TP}{TP+FP}$ measures the accuracy of same cluster word pairs, while recall $R=\frac{TP}{TP+FN}$ measures the ratio of same label word pairs which have been correctly clustered.
We set $\beta=5$ to give more weight to recall by penalizing false negative more strongly. 
\end{itemize}
 
Tables \ref{table:NMI} and \ref{table:Fmeasure} show the clustering evaluation. We can see that our proposed DW2V consistently outperforms other baselines for all values of $K$.
These results show two advantages. 
First, the word semantic shift has been captured by the temporal embeddings (for example, by correlating correctly with the section label of  \texttt{amazon}, which changes from \textit{World} to \textit{Technology}).
Second, since embeddings of words of all years are used for clustering, a good clustering result indicates good alignment across years.
We can also see that AW2V also performs well, as it also applies alignment between adjacent time slices for all words.
However, TW2V does not perform well as others, suggesting that aligning locally (only a few words) is not sufficient for high alignment quality.
% its local alignment strategy for particular words is not a
%by transforming its embedding from one time slot to anther based on $k$ nearest neighbor words. 
%This strategy provides customized alignment for single words, however, it fails to provide a unified alignment for multiple words or whole embedding space.

\subsection{Alignment quality}
%Another important property of temporal word embeddings is that embeddings of static words shuold be consistent over time.
We now more directly evaluate alignment quality, \ie~the property
%Another important property of temporal word embeddings is 
that the semantic distribution in temporal embedding space should be consistent over time.
%the embeddings space for each time slots should be consistent for inter-time comparison.
For example, if a word such as \texttt{estate} or \texttt{republican} does not change much throughout time, its embedding should remain relatively constant for different $t$. 
By the same logic, if a word such as \texttt{trump} does change association throughout time, its embedding should reflect this shift by moving from one position to another
% for associating to different words 
(\eg, \texttt{estate} $\rightarrow$ \texttt{republican}).
We saw this in the previous section for static words like \texttt{president} or \texttt{mayor}; they do not change meanings, though they are accompanied by names that shift to them every few years.
%a static word that is semantically similar to another word at time $t$ should stay similar to its time $t$ embedding for all other $t'$ as well,  \emph{even if the second word is dynamic}.
%As a result, if a new word emerges in later time, once we put its embedding to earlier space, it is still effective to infer the meaning.

\begin{table}[t]
\abovecaptionskip
\belowcaptionskip
%\addtolength{\tabcolsep}{-3pt}
\caption{Mean Reciprocal Rank (MRR) and Mean Precision (MP) for Testset 1.}\label{table:align1}
%\vspace{-0.3cm}
%\scriptsize
%\addtolength{\tabcolsep}{-5pt}
\begin{center}
\small
    \begin{tabular}{  c | c | c | c | c | c}
    \hline\hline
    Method & MRR & MP@1 & MP@3 & MP@5 & MP@10 \\ \hline
    SW2V  & 0.3560 & 0.2664 & 0.4210 & 0.4774 & 0.5612\\
    TW2V  & 0.0920 & 0.0500 & 0.1168 & 0.1482 & 0.1910\\
    AW2V  & 0.1582 & 0.1066 & 0.1814 & 0.2241 & 0.2953\\
   	DW2V & \textbf{0.4222} & \textbf{0.3306} & \textbf{0.4854} & \textbf{0.5488} & \textbf{0.6191}\\
    \hline\hline
    \end{tabular}
\normalsize
\end{center}
%\vspace{-0.3cm}
\end{table}

\begin{table}[t]
\abovecaptionskip
\belowcaptionskip
%\addtolength{\tabcolsep}{-3pt}
\caption{Mean Reciprocal Rank (MRR) and Mean Precision (MP) for Testset 2.}\label{table:align2}
%\vspace{-0.3cm}
%\scriptsize
%\addtolength{\tabcolsep}{-5pt}
\begin{center}
\small
    \begin{tabular}{  c | c | c | c | c | c}
    \hline\hline
    Method & MRR & MP@1 & MP@3 & MP@5 & MP@10 \\ \hline
    SW2V  & 0.0472 & 0.0000 & 0.0787 & 0.0787 & 0.2022\\
    TW2V  & 0.0664 & 0.0404 & 0.0764 & 0.0989 & 0.1438\\
    AW2V  & 0.0500 & 0.0225 & 0.0517 & 0.0787 & 0.1416\\
   	DW2V & \textbf{0.1444} & \textbf{0.0764} & \textbf{0.1596} & \textbf{0.2202} & \textbf{0.3820}\\
    \hline\hline
    \end{tabular}
\normalsize
\end{center}
\vspace{-0.2cm}
\end{table}

To examine the quality of embedding alignment, we create a task to query equivalences across years. 
For example, given  \texttt{obama-2012}, we want to query its equivalent word in 2002. As we know \texttt{obama} is the U.S. president in 2012; its equivalent in 2002 is  \texttt{bush}, who was the U.S. president at that time.
In this way, we create two testsets.
%\footnote{We will make these datasets available upon acceptance.}.

The first one is based on  publicly recorded knowledge that for each year lists different names for a particular role, such as U.S. president, U.K. prime minister, NFL superbowl champion team, and so on.
For each year (\eg, 2012), we put its word (\eg, \texttt{obama}) into the embedding set of every other year for query its equivalence in top closest words. 

The second test is human-generated, for exploring more interesting concepts like emerging technologies, brands and major events (\eg, disease outbreaks and financial crisis).
For constructing the test word pairs, we first select emerging terms which have not been popularized before 1994, then query their well known precedents during 1990 to 1994 (\eg, \texttt{app-2012} can correspond to \texttt{software-1990}).
For emerging word (\eg, app) we extract its most popular year (\eg, 2012) with maximum frequency, and put its embedding into each year from 1990 to 1994 for querying its precedent (\eg, software).
%Compared to the first test, this one is more subjectively judged by human. For example, \texttt{bluray} corresponds to \texttt{dvd} in early time, also, \texttt{google} corresponds to \texttt{ibm} as its precedent.
%We  create these word pairs of cross-time semantically similar words, such as  \texttt{obama-2008}-\texttt{clinton-1992}, \texttt{mp3-2000}-\texttt{cassette-1990}, and so on.
%We do this for presidents of several countries, major tech companies and entities, and several major events such as disease outbreaks and hurricanes.
%For concepts without clear years
%(\eg popularization of BluRay or DVDs) we make educated guesses.
% emerging year and early equivalent year (e.g., \texttt{bluray-2009}-\texttt{dvd-1999}), and put the embeddings of emerging words in labeled years into the embedding space of early equivalent years.
Each word-year pair now forms a query and an answer; in total we have $N = 11028$ such pairs in the first testset, and $N = 445$ in the second one.

%To implement the test, we put the query word's embedding of a particular year into the embedding space of the target year and rank the closest words.
%By querying the emerging word's top-K (e.g., top-20) closed words, we count this case successfully by finding the equivalent, otherwise we count this case failed.

We use two metrics to evaluate the performance.
\begin{itemize} [leftmargin=*]
\item For each test $i$, the correct answer word is identified at position rank$[i]$ for closest words.
The \emph{Mean Reciprocal Rank (MRR)} is defined as
\begin{equation}
\label{MRR}
MRR=\frac{1}{N}\sum_{i=1}^N \frac{1}{\text{rank}[i]},
\end{equation}
where $\frac{1}{\text{rank}[i]} = 0$ if the correct answer is not found in the $top$-10. 
Higher MRR  means that correct answers appear more closely and unambiguously with the  query embedding.
% higher MRR number means the correct answers have higher rank to be discovered.
%where $N$ is the number of testing word pairs and $\text{rank}_i$ refers to the rank position of equivalent word, and 

\item Additionally,  for test $i$ consisting of a query and target word-year pair, 
consider the closest $K$ words to the query embedding in the target year. If the target word is among these $K$ words, then the \emph{Precision@K} for test $i$ (denoted P@K[$i$]) is 1; else, it is 0.
Then the 
\emph{Mean Precision@K} is defined as
\begin{equation}
\label{precision}
MP@K = \frac{1}{N}\sum_{i=1}^N (P@K[i]).
\end{equation}
%where $\mathbb{I}_i$ is the indicator function: 1 for found in top-k while 0 for not found.
Higher precision indicates a better ability to acquire correct answers using close embeddings.
\end{itemize}
%We identify the closest $K$ words to the query embedding in the target year.

Tables \ref{table:align1} and \ref{table:align2} show the evaluation of the alignment test.
We can see that our proposed method outperforms others and shows good alignment quality, sometimes by an order of magnitude.
Comparing to testset 1 which has a large amount of queries with considerable short range alignments (\eg, from 2012 to 2013), the testset 2 mostly consists of fewer long range alignments (\eg, 2012 to 1990).
Therefore, we can see that the performance of SW2V is relatively good in testset 1 since the semantic distribution does not change much in short ranges which makes this test favorable to static embeddings. However, SW2V degrades sharply in testset 2, where the long range alignment is needed more.
For TW2V, since it does an individual year-to-year (\eg, 2012-to-1990) transformation by assuming that the local structure of target words does not shift, its overall alignment quality of whole embedding sets is not satisfied in testset 1 with many alignments. However, it does similarly to AW2V in testset 2 because its individual year-to-year transformation makes it more capable for long range alignment.
AW2V, which enforces alignment for whole embedding sets between adjacent time slices, provides quite reliable performance.
% carries an overall alignment 
%achieves  second place. 
However, 
%due to the separation of embedding learning and alignment operation, 
its alignment quality is still below ours, suggesting that their two-step approach is not as successful in enforcing global alignment. 
%\vspace{-0.2cm}

\subsection{Robustness}
Finally, we explore the robustness of our embedding model against subsampling of words for select years. 
Table \ref{table:robust}  shows the result of the alignment task (testset 1) for vectors computed from subsampled co-occurrence matrices for every three years from 1991 to 2015. 
To subsample, each element $C_{ij}$ is replaced with a randomly drawn integer $\hat C_{ij}$ from a Binomial distribution for rate $r$ and $n=C_{ij}$ trials; this simulates the number of co-occurrences measured if they had been missed with probability $r$. 
The new frequency  $\hat f$ is then renormalized so that  
$\hat f_i/f_i = \sum_j \hat C_{ij} / \sum_j C_{ij}$.
 Listed are the alignment test results for $r=1, 0.1, 0.01,$ and $0.001$ compared against  \cite{hamilton2016diachronic}, which otherwise performs comparably with our embedding. 
Unsurprisingly, for extreme attacks (leaving only 1\% or 0.1\% co-occurrences), the performance of \cite{hamilton2016diachronic} degrades sharply; however, because of our joint optimization approach, the performance of our embeddings seems to hold steady.

\begin{table}%[h]
\abovecaptionskip
\belowcaptionskip
%\addtolength{\tabcolsep}{-3pt}
\caption{MRR and MP for alignment with every 3 years subsampling.}\label{table:robust}
%\vspace{-0.3cm}
%\scriptsize
%\addtolength{\tabcolsep}{-5pt}
\begin{center}
\small
    \begin{tabular}{  c | c | c | c | c | c | c}
    \hline\hline
    Method &$r$  & MRR & MP@1 & MP@3 & MP@5 & MP@10 \\ \hline
    AW2V  & 100\% &  0.1582 & 0.1066 & 0.1814 & 0.2241 & 0.2953\\
    AW2V & 10\%  & 0.0884 & 0.0567 & 0.1020 & 0.1287 & 0.1727\\
    AW2V &1\% & 0.0409 & 0.0255 & 0.0475 & 0.0605 & 0.0818\\
    AW2V & 0.1\% & 0.0362 & 0.0239 & 0.0416 & 0.0532 & 0.0690\\\hline
    DW2V  & 100\%&  0.4222 & 0.3306 & 0.4854 & 0.5488 & 0.6191\\
   	DW2V & 10\%& 0.4394 & 0.3489 & 0.5036 & 0.5628 & 0.6292\\
   	DW2V& 1\% & 0.4418 & 0.3522 & 0.5024 & 0.5636 & 0.6310\\
   	DW2V & 0.1\%& 0.4427 & 0.3550 & 0.5006 & 0.5612 & 0.6299\\
    \hline\hline
    \end{tabular}
\normalsize
\end{center}
\vspace{-0.3cm}
\end{table}

%\vspace{-0.2cm}

\section{Related Work}
%\begin{comment}
\textbf{Temporal effects in natural language processing: }
There are several studies that investigate the temporal features of natural language.
%There are several studies that investigate the use of natural language features, especially word semantic shifts, in data mining field.  
Some are using topic modeling on news corpus 
\cite{allan2001temporal}
or time-stamped scientific journals 
\cite{sipos2012temporal,wang2006topics,blei2006dynamic}
to find spikes and emergences of themes and viewpoints.
Simpler word count features are used in \cite{heyer2009change, michel2011quantitative, choi2012predicting} to find hotly discussed concepts and cultural phenomena, in \cite{merchant2001teenagers, schiano2002teen, tagliamonte2008linguistic} to analyze teen behavior in chatrooms, and in  \cite{heyer2009change} to discover incidents of influenza.
%\end{comment}

\noindent\textbf{Word embedding learning: }
The idea of word embeddings has existed at least since the 90s, with vectors computed as 
rows of the co-occurrence \cite{HAL1996}, 
through matrix factorization \cite{lsa},
and most famously through deep neural networks \cite{bengioneural,collobert2008unified}. 
They have recently been  repopularized with the success of low-dimensional embeddings like GloVE \cite{glove} and 
word2vec \cite{mikolov2013distributed,mikolov2013efficient}, which have been shown to greatly improve the performance in key NLP tasks, like document clustering \cite{kusner2015word}, LDA \cite{petterson2010word}, and word similarity \cite{levy2015improving,baroni2014don}.
There is a close connection between these recent methods and our proposed method, in that both word2vec and GloVE have been shown to be equivalent to matrix factorization of a shifted PMI matrix \cite{levy2014neural}.

\noindent\textbf{Temporal word embeddings and evaluations: }
While NLP tools have been used frequently to discover emerging word meanings and societal trends, many of them rely on changes in the co-occurrence or PMI matrix
\cite{michel2011quantitative,gulordava2011distributional,
wijaya2011understanding,heyer2009change,mitra2014s},
changes in parts of speech,
\cite{mihalcea2012word}
or other statistical methods
\cite{tang2016semantic,basile2014analysing,kulkarni2015statistically}.
A few works use low-dimensional word embeddings, but either do no smoothing \cite{sagi2011tracing}, or use two-step methods
\cite{kim2014temporal, hamilton2016diachronic, kulkarni2015statistically}.
Semantic shift and emergence are also evaluated in many different ways.  In \cite{sagi2011tracing}, word shifts are identified by tracking the mean angle between a word and its neighbors.
One of the several tests in \cite{kulkarni2015statistically} create synthetic data with injected semantic shifts, and quantifies the accuracy of capturing them using various time series metrics. 
In \cite{mihalcea2012word}, the authors show the semantic meaningfulness of key lexical features by using them to predict the time-stamp of a particular phrase.
And, \cite{mitra2014s} makes the connection that emergent meanings usually coexist with previous meanings, and use dynamic embeddings to discover and identify multisenses, evaluated against WordNet.
Primarily, temporal word embeddings are evaluated against human-created databases of known semantically shifted words
\cite{hamilton2016diachronic,kulkarni2015statistically,tang2016semantic} which is our approach as well.
%\vspace{-0.2cm}

%\vspace{-0.2cm}
\section{Conclusion}
\label{sec:conc}
We studied the evolution of word semantics as a dynamic word embedding learning problem. 
We proposed a model to learn time-aware word embeddings and used it to dynamically mine text corpora. 
Our proposed method simultaneously learns the embeddings and aligns them across time, and has several benefits: higher interpretability for embeddings, better quality with less data, and more reliable alignment for across-time querying. We solved the resulting optimization problem using a scalable block coordinate descent method. We designed qualitative and quantitative methods to evaluate temporal embeddings for evolving word semantics, and showed that our dynamic embedding method performs favorably against other temporal embedding approaches.
\bibliographystyle{ACM-Reference-Format}
\bibliography{Ref} 


\begin{thebibliography}{00}

%%% the Web address.

\ifx \showCODEN    \undefined \def \showCODEN     #1{\unskip}     \fi
\ifx \showDOI      \undefined \def \showDOI       #1{{\tt DOI:}\penalty0{#1}\ }
  \fi
\ifx \showISBNx    \undefined \def \showISBNx     #1{\unskip}     \fi
\ifx \showISBNxiii \undefined \def \showISBNxiii  #1{\unskip}     \fi
\ifx \showISSN     \undefined \def \showISSN      #1{\unskip}     \fi
\ifx \showLCCN     \undefined \def \showLCCN      #1{\unskip}     \fi
\ifx \shownote     \undefined \def \shownote      #1{#1}          \fi
\ifx \showarticletitle \undefined \def \showarticletitle #1{#1}   \fi
\ifx \showURL      \undefined \def \showURL       #1{#1}          \fi
% invisible to TeX
\providecommand\bibfield[2]{#2}
\providecommand\bibinfo[2]{#2}
\providecommand\natexlab[1]{#1}
\providecommand\showeprint[2][]{arXiv:#2}

\bibitem[\protect\citeauthoryear{Allan, Gupta, and Khandelwal}{Allan
  et~al\mbox{.}}{2001}]%
        {allan2001temporal}
\bibfield{author}{\bibinfo{person}{James Allan}, \bibinfo{person}{Rahul Gupta},
  {and} \bibinfo{person}{Vikas Khandelwal}.} \bibinfo{year}{2001}\natexlab{}.
\newblock \showarticletitle{Temporal summaries of new topics}. In
  \bibinfo{booktitle}{{\em Proceedings of the 24th annual international ACM
  SIGIR conference on Research and development in information retrieval}}. ACM,
  \bibinfo{pages}{10--18}.
\newblock


\bibitem[\protect\citeauthoryear{Arora, Li, Liang, Ma, and Risteski}{Arora
  et~al\mbox{.}}{2015}]%
        {arora2015rand}
\bibfield{author}{\bibinfo{person}{Sanjeev Arora}, \bibinfo{person}{Yuanzhi
  Li}, \bibinfo{person}{Yingyu Liang}, \bibinfo{person}{Tengyu Ma}, {and}
  \bibinfo{person}{Andrej Risteski}.} \bibinfo{year}{2015}\natexlab{}.
\newblock \showarticletitle{Rand-walk: A latent variable model approach to word
  embeddings}.
\newblock \bibinfo{journal}{{\em arXiv preprint arXiv:1502.03520\/}}
  (\bibinfo{year}{2015}).
\newblock


\bibitem[\protect\citeauthoryear{Baroni, Dinu, and Kruszewski}{Baroni
  et~al\mbox{.}}{2014}]%
        {baroni2014don}
\bibfield{author}{\bibinfo{person}{Marco Baroni}, \bibinfo{person}{Georgiana
  Dinu}, {and} \bibinfo{person}{Germ{\'a}n Kruszewski}.}
  \bibinfo{year}{2014}\natexlab{}.
\newblock \showarticletitle{Don't count, predict! A systematic comparison of
  context-counting vs. context-predicting semantic vectors.}. In
  \bibinfo{booktitle}{{\em ACL (1)}}. \bibinfo{pages}{238--247}.
\newblock


\bibitem[\protect\citeauthoryear{Basile, Caputo, and Semeraro}{Basile
  et~al\mbox{.}}{2014}]%
        {basile2014analysing}
\bibfield{author}{\bibinfo{person}{Pierpaolo Basile}, \bibinfo{person}{Annalina
  Caputo}, {and} \bibinfo{person}{Giovanni Semeraro}.}
  \bibinfo{year}{2014}\natexlab{}.
\newblock \showarticletitle{Analysing word meaning over time by exploiting
  temporal random indexing}. In \bibinfo{booktitle}{{\em First Italian
  Conference on Computational Linguistics CLiC-it}}.
\newblock


\bibitem[\protect\citeauthoryear{Bengio, Ducharme, Vincent, and Jauvin}{Bengio
  et~al\mbox{.}}{2003}]%
        {bengioneural}
\bibfield{author}{\bibinfo{person}{Yoshua Bengio}, \bibinfo{person}{R{\'e}jean
  Ducharme}, \bibinfo{person}{Pascal Vincent}, {and} \bibinfo{person}{Christian
  Jauvin}.} \bibinfo{year}{2003}\natexlab{}.
\newblock \showarticletitle{A Neural Probabilistic Language Model}.
\newblock \bibinfo{journal}{{\em Journal of Machine Learning Research\/}}
  \bibinfo{volume}{3} (\bibinfo{year}{2003}), \bibinfo{pages}{1137--1155}.
\newblock


\bibitem[\protect\citeauthoryear{Blei and Lafferty}{Blei and Lafferty}{2006}]%
        {blei2006dynamic}
\bibfield{author}{\bibinfo{person}{David~M Blei} {and} \bibinfo{person}{John~D
  Lafferty}.} \bibinfo{year}{2006}\natexlab{}.
\newblock \showarticletitle{Dynamic topic models}. In \bibinfo{booktitle}{{\em
  Proceedings of the 23rd international conference on Machine learning}}. ACM,
  \bibinfo{pages}{113--120}.
\newblock


\bibitem[\protect\citeauthoryear{Choi and Varian}{Choi and Varian}{2012}]%
        {choi2012predicting}
\bibfield{author}{\bibinfo{person}{Hyunyoung Choi} {and} \bibinfo{person}{Hal
  Varian}.} \bibinfo{year}{2012}\natexlab{}.
\newblock \showarticletitle{Predicting the present with Google Trends}.
\newblock \bibinfo{journal}{{\em Economic Record\/}} \bibinfo{volume}{88},
  \bibinfo{number}{s1} (\bibinfo{year}{2012}), \bibinfo{pages}{2--9}.
\newblock


\bibitem[\protect\citeauthoryear{Collobert and Weston}{Collobert and
  Weston}{2008}]%
        {collobert2008unified}
\bibfield{author}{\bibinfo{person}{Ronan Collobert} {and}
  \bibinfo{person}{Jason Weston}.} \bibinfo{year}{2008}\natexlab{}.
\newblock \showarticletitle{A unified architecture for natural language
  processing: Deep neural networks with multitask learning}. In
  \bibinfo{booktitle}{{\em Proceedings of the 25th international conference on
  Machine learning}}. ACM, \bibinfo{pages}{160--167}.
\newblock


\bibitem[\protect\citeauthoryear{Deerwester, Dumais, Furnas, Landauer, and
  Harshman}{Deerwester et~al\mbox{.}}{1990}]%
        {lsa}
\bibfield{author}{\bibinfo{person}{Scott Deerwester}, \bibinfo{person}{Susan~T
  Dumais}, \bibinfo{person}{George~W Furnas}, \bibinfo{person}{Thomas~K
  Landauer}, {and} \bibinfo{person}{Richard Harshman}.}
  \bibinfo{year}{1990}\natexlab{}.
\newblock \showarticletitle{Indexing by latent semantic analysis}.
\newblock \bibinfo{journal}{{\em Journal of the American society for
  information science\/}} \bibinfo{volume}{41}, \bibinfo{number}{6}
  (\bibinfo{year}{1990}), \bibinfo{pages}{391}.
\newblock


\bibitem[\protect\citeauthoryear{Firth}{Firth}{1957}]%
        {firth1957synopsis}
\bibfield{author}{\bibinfo{person}{John~R Firth}.}
  \bibinfo{year}{1957}\natexlab{}.
\newblock \showarticletitle{$\{$A synopsis of linguistic theory,
  1930-1955$\}$}.
\newblock  (\bibinfo{year}{1957}).
\newblock


\bibitem[\protect\citeauthoryear{Gulordava and Baroni}{Gulordava and
  Baroni}{2011}]%
        {gulordava2011distributional}
\bibfield{author}{\bibinfo{person}{Kristina Gulordava} {and}
  \bibinfo{person}{Marco Baroni}.} \bibinfo{year}{2011}\natexlab{}.
\newblock \showarticletitle{A distributional similarity approach to the
  detection of semantic change in the Google Books Ngram corpus}. In
  \bibinfo{booktitle}{{\em Proceedings of the GEMS 2011 Workshop on GEometrical
  Models of Natural Language Semantics}}. Association for Computational
  Linguistics, \bibinfo{pages}{67--71}.
\newblock


\bibitem[\protect\citeauthoryear{Hamilton, Leskovec, and Jurafsky}{Hamilton
  et~al\mbox{.}}{2016}]%
        {hamilton2016diachronic}
\bibfield{author}{\bibinfo{person}{William~L Hamilton}, \bibinfo{person}{Jure
  Leskovec}, {and} \bibinfo{person}{Dan Jurafsky}.}
  \bibinfo{year}{2016}\natexlab{}.
\newblock \showarticletitle{Diachronic Word Embeddings Reveal Statistical Laws
  of Semantic Change}.
\newblock \bibinfo{journal}{{\em arXiv preprint arXiv:1605.09096\/}}
  (\bibinfo{year}{2016}).
\newblock


\bibitem[\protect\citeauthoryear{Heyer, Holz, and Teresniak}{Heyer
  et~al\mbox{.}}{2009}]%
        {heyer2009change}
\bibfield{author}{\bibinfo{person}{Gerhard Heyer}, \bibinfo{person}{Florian
  Holz}, {and} \bibinfo{person}{Sven Teresniak}.}
  \bibinfo{year}{2009}\natexlab{}.
\newblock \showarticletitle{Change of Topics over Time-Tracking Topics by their
  Change of Meaning.}
\newblock \bibinfo{journal}{{\em KDIR\/}}  \bibinfo{volume}{9}
  (\bibinfo{year}{2009}), \bibinfo{pages}{223--228}.
\newblock


\bibitem[\protect\citeauthoryear{Kim, Chiu, Hanaki, Hegde, and Petrov}{Kim
  et~al\mbox{.}}{2014}]%
        {kim2014temporal}
\bibfield{author}{\bibinfo{person}{Yoon Kim}, \bibinfo{person}{Yi-I Chiu},
  \bibinfo{person}{Kentaro Hanaki}, \bibinfo{person}{Darshan Hegde}, {and}
  \bibinfo{person}{Slav Petrov}.} \bibinfo{year}{2014}\natexlab{}.
\newblock \showarticletitle{Temporal analysis of language through neural
  language models}.
\newblock \bibinfo{journal}{{\em arXiv preprint arXiv:1405.3515\/}}
  (\bibinfo{year}{2014}).
\newblock


\bibitem[\protect\citeauthoryear{Kulkarni, Al-Rfou, Perozzi, and
  Skiena}{Kulkarni et~al\mbox{.}}{2015}]%
        {kulkarni2015statistically}
\bibfield{author}{\bibinfo{person}{Vivek Kulkarni}, \bibinfo{person}{Rami
  Al-Rfou}, \bibinfo{person}{Bryan Perozzi}, {and} \bibinfo{person}{Steven
  Skiena}.} \bibinfo{year}{2015}\natexlab{}.
\newblock \showarticletitle{Statistically significant detection of linguistic
  change}. In \bibinfo{booktitle}{{\em Proceedings of the 24th International
  Conference on World Wide Web}}. ACM, \bibinfo{pages}{625--635}.
\newblock


\bibitem[\protect\citeauthoryear{Kusner, Sun, Kolkin, Weinberger,
  et~al\mbox{.}}{Kusner et~al\mbox{.}}{2015}]%
        {kusner2015word}
\bibfield{author}{\bibinfo{person}{Matt~J Kusner}, \bibinfo{person}{Yu Sun},
  \bibinfo{person}{Nicholas~I Kolkin}, \bibinfo{person}{Kilian~Q Weinberger},
  {and} \bibinfo{person}{others}.} \bibinfo{year}{2015}\natexlab{}.
\newblock \showarticletitle{From Word Embeddings To Document Distances.}. In
  \bibinfo{booktitle}{{\em ICML}}, Vol.~\bibinfo{volume}{15}.
  \bibinfo{pages}{957--966}.
\newblock


\bibitem[\protect\citeauthoryear{Levy and Goldberg}{Levy and Goldberg}{2014}]%
        {levy2014neural}
\bibfield{author}{\bibinfo{person}{Omer Levy} {and} \bibinfo{person}{Yoav
  Goldberg}.} \bibinfo{year}{2014}\natexlab{}.
\newblock \showarticletitle{Neural word embedding as implicit matrix
  factorization}. In \bibinfo{booktitle}{{\em Advances in neural information
  processing systems}}. \bibinfo{pages}{2177--2185}.
\newblock


\bibitem[\protect\citeauthoryear{Levy, Goldberg, and Dagan}{Levy
  et~al\mbox{.}}{2015}]%
        {levy2015improving}
\bibfield{author}{\bibinfo{person}{Omer Levy}, \bibinfo{person}{Yoav Goldberg},
  {and} \bibinfo{person}{Ido Dagan}.} \bibinfo{year}{2015}\natexlab{}.
\newblock \showarticletitle{Improving distributional similarity with lessons
  learned from word embeddings}.
\newblock \bibinfo{journal}{{\em Transactions of the Association for
  Computational Linguistics\/}}  \bibinfo{volume}{3} (\bibinfo{year}{2015}),
  \bibinfo{pages}{211--225}.
\newblock


\bibitem[\protect\citeauthoryear{Liao and Cheng}{Liao and Cheng}{2016}]%
        {liao2016analysing}
\bibfield{author}{\bibinfo{person}{Xuanyi Liao} {and} \bibinfo{person}{Guang
  Cheng}.} \bibinfo{year}{2016}\natexlab{}.
\newblock \showarticletitle{Analysing the Semantic Change Based on Word
  Embedding}. In \bibinfo{booktitle}{{\em International Conference on Computer
  Processing of Oriental Languages}}. Springer, \bibinfo{pages}{213--223}.
\newblock


\bibitem[\protect\citeauthoryear{Lund and Burgess}{Lund and Burgess}{1996}]%
        {HAL1996}
\bibfield{author}{\bibinfo{person}{Kevin Lund} {and} \bibinfo{person}{Curt
  Burgess}.} \bibinfo{year}{1996}\natexlab{}.
\newblock \showarticletitle{Producing high-dimensional semantic spaces from
  lexical co-occurrence}.
\newblock \bibinfo{journal}{{\em Behavior Research Methods, Instruments, \&
  Computers\/}} \bibinfo{volume}{28}, \bibinfo{number}{2}
  (\bibinfo{year}{1996}), \bibinfo{pages}{203--208}.
\newblock


\bibitem[\protect\citeauthoryear{Merchant}{Merchant}{2001}]%
        {merchant2001teenagers}
\bibfield{author}{\bibinfo{person}{Guy Merchant}.}
  \bibinfo{year}{2001}\natexlab{}.
\newblock \showarticletitle{Teenagers in cyberspace: An investigation of
  language use and language change in internet chatrooms}.
\newblock \bibinfo{journal}{{\em Journal of Research in Reading\/}}
  \bibinfo{volume}{24}, \bibinfo{number}{3} (\bibinfo{year}{2001}),
  \bibinfo{pages}{293--306}.
\newblock


\bibitem[\protect\citeauthoryear{Michel, Shen, Aiden, Veres, Gray, Pickett,
  Hoiberg, Clancy, Norvig, Orwant, et~al\mbox{.}}{Michel et~al\mbox{.}}{2011}]%
        {michel2011quantitative}
\bibfield{author}{\bibinfo{person}{Jean-Baptiste Michel},
  \bibinfo{person}{Yuan~Kui Shen}, \bibinfo{person}{Aviva~Presser Aiden},
  \bibinfo{person}{Adrian Veres}, \bibinfo{person}{Matthew~K Gray},
  \bibinfo{person}{Joseph~P Pickett}, \bibinfo{person}{Dale Hoiberg},
  \bibinfo{person}{Dan Clancy}, \bibinfo{person}{Peter Norvig},
  \bibinfo{person}{Jon Orwant}, {and} \bibinfo{person}{others}.}
  \bibinfo{year}{2011}\natexlab{}.
\newblock \showarticletitle{Quantitative analysis of culture using millions of
  digitized books}.
\newblock \bibinfo{journal}{{\em science\/}} \bibinfo{volume}{331},
  \bibinfo{number}{6014} (\bibinfo{year}{2011}), \bibinfo{pages}{176--182}.
\newblock


\bibitem[\protect\citeauthoryear{Mihalcea and Nastase}{Mihalcea and
  Nastase}{2012}]%
        {mihalcea2012word}
\bibfield{author}{\bibinfo{person}{Rada Mihalcea} {and} \bibinfo{person}{Vivi
  Nastase}.} \bibinfo{year}{2012}\natexlab{}.
\newblock \showarticletitle{Word epoch disambiguation: Finding how words change
  over time}. In \bibinfo{booktitle}{{\em Proceedings of the 50th Annual
  Meeting of the Association for Computational Linguistics: Short Papers-Volume
  2}}. Association for Computational Linguistics, \bibinfo{pages}{259--263}.
\newblock


\bibitem[\protect\citeauthoryear{Mikolov, Chen, Corrado, and Dean}{Mikolov
  et~al\mbox{.}}{2013a}]%
        {mikolov2013efficient}
\bibfield{author}{\bibinfo{person}{Tomas Mikolov}, \bibinfo{person}{Kai Chen},
  \bibinfo{person}{Greg Corrado}, {and} \bibinfo{person}{Jeffrey Dean}.}
  \bibinfo{year}{2013}\natexlab{a}.
\newblock \showarticletitle{Efficient estimation of word representations in
  vector space}.
\newblock \bibinfo{journal}{{\em arXiv preprint arXiv:1301.3781\/}}
  (\bibinfo{year}{2013}).
\newblock


\bibitem[\protect\citeauthoryear{Mikolov, Sutskever, Chen, Corrado, and
  Dean}{Mikolov et~al\mbox{.}}{2013b}]%
        {mikolov2013distributed}
\bibfield{author}{\bibinfo{person}{Tomas Mikolov}, \bibinfo{person}{Ilya
  Sutskever}, \bibinfo{person}{Kai Chen}, \bibinfo{person}{Greg~S Corrado},
  {and} \bibinfo{person}{Jeff Dean}.} \bibinfo{year}{2013}\natexlab{b}.
\newblock \showarticletitle{Distributed representations of words and phrases
  and their compositionality}. In \bibinfo{booktitle}{{\em Advances in neural
  information processing systems}}. \bibinfo{pages}{3111--3119}.
\newblock


\bibitem[\protect\citeauthoryear{Mitra, Mitra, Riedl, Biemann, Mukherjee, and
  Goyal}{Mitra et~al\mbox{.}}{2014}]%
        {mitra2014s}
\bibfield{author}{\bibinfo{person}{Sunny Mitra}, \bibinfo{person}{Ritwik
  Mitra}, \bibinfo{person}{Martin Riedl}, \bibinfo{person}{Chris Biemann},
  \bibinfo{person}{Animesh Mukherjee}, {and} \bibinfo{person}{Pawan Goyal}.}
  \bibinfo{year}{2014}\natexlab{}.
\newblock \showarticletitle{That's sick dude!: Automatic identification of word
  sense change across different timescales}.
\newblock \bibinfo{journal}{{\em arXiv preprint arXiv:1405.4392\/}}
  (\bibinfo{year}{2014}).
\newblock


\bibitem[\protect\citeauthoryear{Pennington, Socher, and Manning}{Pennington
  et~al\mbox{.}}{2014}]%
        {glove}
\bibfield{author}{\bibinfo{person}{Jeffrey Pennington},
  \bibinfo{person}{Richard Socher}, {and} \bibinfo{person}{Christopher~D
  Manning}.} \bibinfo{year}{2014}\natexlab{}.
\newblock \showarticletitle{Glove: Global Vectors for Word Representation.}. In
  \bibinfo{booktitle}{{\em EMNLP}}, Vol.~\bibinfo{volume}{14}.
  \bibinfo{pages}{1532--1543}.
\newblock


\bibitem[\protect\citeauthoryear{Petterson, Buntine, Narayanamurthy, Caetano,
  and Smola}{Petterson et~al\mbox{.}}{2010}]%
        {petterson2010word}
\bibfield{author}{\bibinfo{person}{James Petterson}, \bibinfo{person}{Wray
  Buntine}, \bibinfo{person}{Shravan~M Narayanamurthy},
  \bibinfo{person}{Tib{\'e}rio~S Caetano}, {and} \bibinfo{person}{Alex~J
  Smola}.} \bibinfo{year}{2010}\natexlab{}.
\newblock \showarticletitle{Word features for latent dirichlet allocation}. In
  \bibinfo{booktitle}{{\em Advances in Neural Information Processing Systems}}.
  \bibinfo{pages}{1921--1929}.
\newblock


\bibitem[\protect\citeauthoryear{Powell}{Powell}{1973}]%
        {powell1973search}
\bibfield{author}{\bibinfo{person}{Michael~JD Powell}.}
  \bibinfo{year}{1973}\natexlab{}.
\newblock \showarticletitle{On search directions for minimization algorithms}.
\newblock \bibinfo{journal}{{\em Mathematical Programming\/}}
  \bibinfo{volume}{4}, \bibinfo{number}{1} (\bibinfo{year}{1973}),
  \bibinfo{pages}{193--201}.
\newblock


\bibitem[\protect\citeauthoryear{Rao, Yu, Ravikumar, and Dhillon}{Rao
  et~al\mbox{.}}{2015}]%
        {rao2015collaborative}
\bibfield{author}{\bibinfo{person}{Nikhil Rao}, \bibinfo{person}{Hsiang-Fu Yu},
  \bibinfo{person}{Pradeep~K Ravikumar}, {and} \bibinfo{person}{Inderjit~S
  Dhillon}.} \bibinfo{year}{2015}\natexlab{}.
\newblock \showarticletitle{Collaborative filtering with graph information:
  Consistency and scalable methods}. In \bibinfo{booktitle}{{\em Advances in
  neural information processing systems}}. \bibinfo{pages}{2107--2115}.
\newblock


\bibitem[\protect\citeauthoryear{Sagi, Kaufmann, and Clark}{Sagi
  et~al\mbox{.}}{2011}]%
        {sagi2011tracing}
\bibfield{author}{\bibinfo{person}{Eyal Sagi}, \bibinfo{person}{Stefan
  Kaufmann}, {and} \bibinfo{person}{Brady Clark}.}
  \bibinfo{year}{2011}\natexlab{}.
\newblock \showarticletitle{Tracing semantic change with latent semantic
  analysis}.
\newblock \bibinfo{journal}{{\em Current methods in historical semantics\/}}
  (\bibinfo{year}{2011}), \bibinfo{pages}{161--183}.
\newblock


\bibitem[\protect\citeauthoryear{Schiano, Chen, Isaacs, Ginsberg,
  Gretarsdottir, and Huddleston}{Schiano et~al\mbox{.}}{2002}]%
        {schiano2002teen}
\bibfield{author}{\bibinfo{person}{Diane~J Schiano}, \bibinfo{person}{Coreena~P
  Chen}, \bibinfo{person}{Ellen Isaacs}, \bibinfo{person}{Jeremy Ginsberg},
  \bibinfo{person}{Unnur Gretarsdottir}, {and} \bibinfo{person}{Megan
  Huddleston}.} \bibinfo{year}{2002}\natexlab{}.
\newblock \showarticletitle{Teen use of messaging media}. In
  \bibinfo{booktitle}{{\em CHI'02 extended abstracts on Human factors in
  computing systems}}. ACM, \bibinfo{pages}{594--595}.
\newblock


\bibitem[\protect\citeauthoryear{Sipos, Swaminathan, Shivaswamy, and
  Joachims}{Sipos et~al\mbox{.}}{2012}]%
        {sipos2012temporal}
\bibfield{author}{\bibinfo{person}{Ruben Sipos}, \bibinfo{person}{Adith
  Swaminathan}, \bibinfo{person}{Pannaga Shivaswamy}, {and}
  \bibinfo{person}{Thorsten Joachims}.} \bibinfo{year}{2012}\natexlab{}.
\newblock \showarticletitle{Temporal corpus summarization using submodular word
  coverage}. In \bibinfo{booktitle}{{\em Proceedings of the 21st ACM
  international conference on Information and knowledge management}}. ACM,
  \bibinfo{pages}{754--763}.
\newblock


\bibitem[\protect\citeauthoryear{Tagliamonte and Denis}{Tagliamonte and
  Denis}{2008}]%
        {tagliamonte2008linguistic}
\bibfield{author}{\bibinfo{person}{Sali~A Tagliamonte} {and}
  \bibinfo{person}{Derek Denis}.} \bibinfo{year}{2008}\natexlab{}.
\newblock \showarticletitle{Linguistic ruin? LOL! Instant messaging and teen
  language}.
\newblock \bibinfo{journal}{{\em American speech\/}} \bibinfo{volume}{83},
  \bibinfo{number}{1} (\bibinfo{year}{2008}), \bibinfo{pages}{3--34}.
\newblock


\bibitem[\protect\citeauthoryear{Tang, Qu, and Chen}{Tang
  et~al\mbox{.}}{2016}]%
        {tang2016semantic}
\bibfield{author}{\bibinfo{person}{Xuri Tang}, \bibinfo{person}{Weiguang Qu},
  {and} \bibinfo{person}{Xiaohe Chen}.} \bibinfo{year}{2016}\natexlab{}.
\newblock \showarticletitle{Semantic change computation: A successive
  approach}.
\newblock \bibinfo{journal}{{\em World Wide Web\/}} \bibinfo{volume}{19},
  \bibinfo{number}{3} (\bibinfo{year}{2016}), \bibinfo{pages}{375--415}.
\newblock


\bibitem[\protect\citeauthoryear{Wang and McCallum}{Wang and McCallum}{2006}]%
        {wang2006topics}
\bibfield{author}{\bibinfo{person}{Xuerui Wang} {and} \bibinfo{person}{Andrew
  McCallum}.} \bibinfo{year}{2006}\natexlab{}.
\newblock \showarticletitle{Topics over time: a non-Markov continuous-time
  model of topical trends}. In \bibinfo{booktitle}{{\em Proceedings of the 12th
  ACM SIGKDD international conference on Knowledge discovery and data mining}}.
  ACM, \bibinfo{pages}{424--433}.
\newblock


\bibitem[\protect\citeauthoryear{Wijaya and Yeniterzi}{Wijaya and
  Yeniterzi}{2011}]%
        {wijaya2011understanding}
\bibfield{author}{\bibinfo{person}{Derry~Tanti Wijaya} {and}
  \bibinfo{person}{Reyyan Yeniterzi}.} \bibinfo{year}{2011}\natexlab{}.
\newblock \showarticletitle{Understanding semantic change of words over
  centuries}. In \bibinfo{booktitle}{{\em Proceedings of the 2011 international
  workshop on DETecting and Exploiting Cultural diversiTy on the social web}}.
  ACM, \bibinfo{pages}{35--40}.
\newblock


\bibitem[\protect\citeauthoryear{Wright}{Wright}{2015}]%
        {wright2015coordinate}
\bibfield{author}{\bibinfo{person}{Stephen~J Wright}.}
  \bibinfo{year}{2015}\natexlab{}.
\newblock \showarticletitle{Coordinate descent algorithms}.
\newblock \bibinfo{journal}{{\em Mathematical Programming\/}}
  \bibinfo{volume}{151}, \bibinfo{number}{1} (\bibinfo{year}{2015}),
  \bibinfo{pages}{3--34}.
\newblock


\bibitem[\protect\citeauthoryear{Yu, Hsieh, Si, and Dhillon}{Yu
  et~al\mbox{.}}{2012}]%
        {yu2012scalable}
\bibfield{author}{\bibinfo{person}{Hsiang-Fu Yu}, \bibinfo{person}{Cho-Jui
  Hsieh}, \bibinfo{person}{Si Si}, {and} \bibinfo{person}{Inderjit Dhillon}.}
  \bibinfo{year}{2012}\natexlab{}.
\newblock \showarticletitle{Scalable coordinate descent approaches to parallel
  matrix factorization for recommender systems}. In \bibinfo{booktitle}{{\em
  12th IEEE International Conference on Data Mining (ICDM)}}. IEEE,
  \bibinfo{pages}{765--774}.
\newblock


\bibitem[\protect\citeauthoryear{Zhang, Jatowt, Bhowmick, and Tanaka}{Zhang
  et~al\mbox{.}}{2016}]%
        {zhang2016past}
\bibfield{author}{\bibinfo{person}{Yating Zhang}, \bibinfo{person}{Adam
  Jatowt}, \bibinfo{person}{Sourav~S Bhowmick}, {and} \bibinfo{person}{Katsumi
  Tanaka}.} \bibinfo{year}{2016}\natexlab{}.
\newblock \showarticletitle{The Past is Not a Foreign Country: Detecting
  Semantically Similar Terms across Time}.
\newblock \bibinfo{journal}{{\em IEEE Transactions on Knowledge and Data
  Engineering\/}} \bibinfo{volume}{28}, \bibinfo{number}{10}
  (\bibinfo{year}{2016}), \bibinfo{pages}{2793--2807}.
\newblock


\end{thebibliography}

\end{document}